\documentclass[fleqn,10pt]{wlscirep}

\usepackage{wrapfig}

\usepackage{graphicx}%
\usepackage{multirow}%
\usepackage{amsmath,amssymb,amsfonts}%
\usepackage{amsthm}%
\usepackage{mathrsfs}%
\usepackage[title]{appendix}%
\usepackage{xcolor}%
\usepackage{textcomp}%
\usepackage{manyfoot}%
\usepackage{booktabs}%
\usepackage{algorithm}%
\usepackage{algorithmicx}%
\usepackage{algpseudocode}%
\usepackage{listings}%
\usepackage{caption}%
\usepackage{subcaption}%
\usepackage[disable]{todonotes}
\usepackage{tikz}

\usepackage[most]{tcolorbox} 
\usepackage{xcolor}          

\theoremstyle{thmstyleone}%
\theoremstyle{thmstyletwo}%

\theoremstyle{thmstylethree}%

\begin{document}

\title{Smart Cellular Bricks for Decentralized Shape Classification and Damage Recovery}

\author[1,*]{Rodrigo Moreno}

\author[1,*]{Andres Faina}

\author[2]{Shyam Sudhakaran}

\author[1]{Kathryn Walker}

\author[1,3,+]{Sebastian Risi}

\affil[1]{IT University Copenhagen, Rued Langgaards Vej 7, Copenhagen, 2300, Denmark}
\affil[2]{Autodesk Research}
\affil[3]{Sakana AI}
\affil[*]{These authors contributed equally.}
\affil[+]{email: sebr@itu.dk}


\begin{abstract}
Biological systems possess remarkable capabilities for self-recognition and morphological regeneration, often relying solely on local interactions. Inspired by these decentralized processes, we present a novel system of physical 3D bricks—simple cubic units equipped with local communication, processing, and sensing—that are capable of inferring their global shape class and detecting structural damage. Leveraging Neural Cellular Automata (NCA), a learned, fully-distributed algorithm, our system enables each module to independently execute the same neural network without access to any global state or positioning information. We demonstrate the ability of collections of hundreds of these cellular bricks to accurately classify a variety of 3D shapes through purely local interactions. The approach shows strong robustness to out-of-distribution shape variations and high tolerance to communication faults and failed modules. In addition to shape inference, the same decentralized framework is extended to detect missing or damaged components, allowing the collective to localize structural disruptions and to guide a recovery process.  This work provides a physical realization of large-scale, decentralized self-recognition and damage detection, advancing the potential of robust, adaptive, and bio-inspired modular systems. Videos and code will be made available at: \url{cellularbricks.github.io/}. 
\end{abstract}




\maketitle


\section*{INTRODUCTION}

Many biological systems exhibit a remarkable capacity to accurately determine their anatomical structure. Through local communication and self-organization, groups of cells can assess whether they have correctly formed a target shape, such as an organ. Moreover, they can actively remodel body parts following injury. For instance, a salamander can regenerate a damaged tail that transforms into a functional leg \cite{farinella1956transformation, vieira2020advancements}, and simple organisms like Hydra and Planaria can fully restore their morphology, regardless of which part is lost \cite{levin2019endogenous, vogg2019model}. This ability to classify general anatomical features—rather than matching a fixed target shape—enables variability among individuals, enhancing the robustness of the process. While the overall function and design of an organ may be consistent across a species, its specific shape, size, or scale can differ from one individual to another.

Artificial systems composed of many physically distributed modules that can autonomously infer their structural class — without centralized control — would represent a significant step toward more adaptable, intelligent artificial  collectives. Such systems could enable powerful applications in smart materials and reconfigurable robotics, where global knowledge must emerge from local sensing and communication.

Closely related to ability to determining ones shape  is the concept of self-assembly, a longstanding goal in modular robotics, pursued for nearly three decades. This effort has led to the development of various systems capable of autonomously assembling into predefined target shapes \cite{stoy2010self, stoy2010modular, abdel2022self, murata2007self,jorgensen2004modular}. However, most of these systems fall short in generalizing to new shapes or detecting damage in a robust, distributed manner. They often rely on centralized computation, manually designed behaviors, or extensive communication protocols that do not scale well. Here we contend that a reliable realization of self-assembly has remained elusive, in part because shape inference is a critical missing ingredient. 

Biological systems offer a promising blueprint. Organisms like Hydra and Planaria exhibit extraordinary self-organizing behavior, capable of robust shape classification and full-body morphological repair through distributed local interactions\cite{levin2019endogenous, vogg2019model}. Even more complex animals like salamanders demonstrate localized, decentralized responses to damage that enable them to infer missing structures based on general shape categories rather than fixed templates. However, existing  prior trying to follow this blueprint either exists only in simulation \cite{randazzo2020self} or in small 2D hardware implementations \cite{walker2022physical}, limiting its applicability to real-world  systems with hundreds or even thousands of components.

Motivated by the scalability and resilience of the collective intelligence of biological systems \cite{mcmillen2024collective}, we introduce a fully decentralized system in which hundreds of physically embodied ``cellular'' bricks  collectively classify their global shape and detect local damage. Our system differs fundamentally from prior  self-modeling robotics research by achieving distributed morphological understanding without relying on actuation \cite{bongard2006resilient,chen2022fully,cully2015robots}, or explicit rule-based coordination \cite{rubenstein2014programmable,kirby2011blinky}. Our cellular bricks achieve a form of emergent self-recognition without movement, sensors, or global model reconstruction. More closely related work by Slavkov et al.~\cite{slavkov2018morphogenesis} focuses on external physical morphogenesis, but it lacks the ability to self-classify the resulting structures—a limitation that becomes critical if the structures are externally modified or not built perfectly.

\begin{figure}
     \centering
     \includegraphics[width=\textwidth]{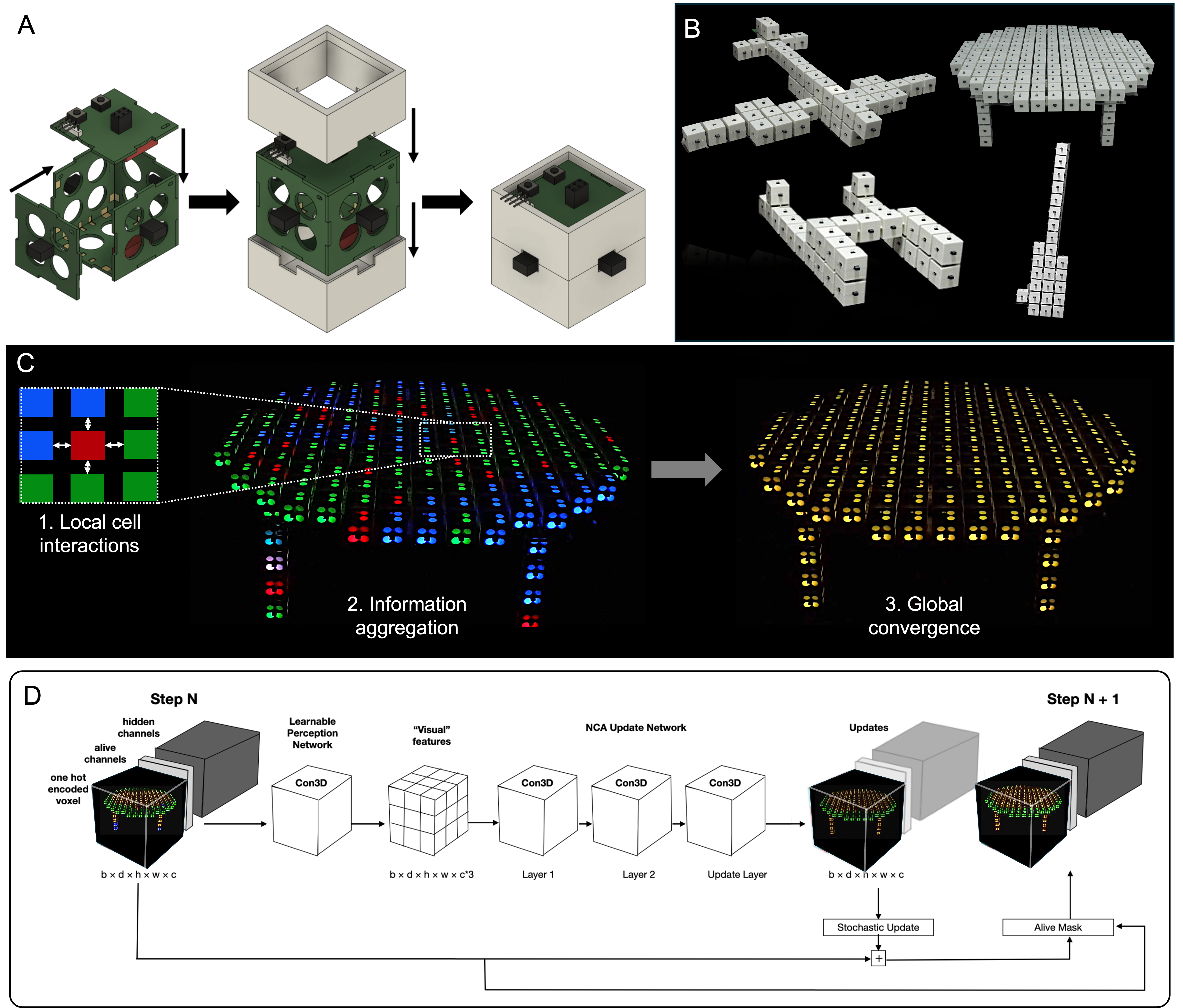}
     \caption{\textbf{Neural cellular automata for shape classification}. \normalfont (\textbf{A}) Cellular brick module. From left to right: Six FR4 PCB faces attached together to form a cube. The PCBs house a microcontroller unit to run the NN and a LED that displays the classification result. All faces contain traces that route power and communication to the attached neighbors through the pin header connectors. PCBs are enclosed in a cover that diffuses the LED light. (\textbf{B}) Examples of cellular bricks assembled into four different shapes. (\textbf{C}) Cells   takes as input local information from their connecting neighboring cells and their hidden channels. Information is aggregated locally, enabling the object to recognize its particular shape over multiple iterations. (\textbf{D}) The local update rules are encoded with a neural cellular automata, a deep neural network. Cells use a 3D convolution with a biologically inspired cross-shaped kernel to gather local information. A two-layer state update module then computes changes to the cell's state based on activation thresholds and stochastic firing, promoting robustness and coordination. The state of a cell is defined by a one-hot encoding of the object class and an additional 20 hidden channels.}
     \label{fig:overview}
 \end{figure}

The collective intelligence algorithm we developed for shape classification is built on the framework of Neural Cellular Automata (NCA) 
 \cite{sudhakaran2021growing,mordvintsev2020growing,wulff1992learning,nichele2017neat,randazzo2020self,hartl2024evolutionary}, extended to operate in 3D and implemented on physical modular hardware. 
Unlike traditional cellular automata (CA) that operate with discrete cell states and hand-crafted rules, NCAs use continuous-valued cell states, enabling end-to-end differentiability and compatibility with gradient descent–based learning algorithms \cite{sudhakaran2021growing,mordvintsev2020growing}. On a high level, each cell in our system is tasked to determine which type of shape it is a part of, solely based on communication with its local neighbors and its memory state. The update rules are parameterized by a deep neural network,  consisting of 3D convolutional layers -- where the network outputs updates that are added to the state of each cell. Each cell state is represented by  28-dimensional state vector, composed of one alpha channel (used to determine if a cell is considered ``dead'' or ''alive), 20 hidden channels that serve as internal memory for each cell, and seven channels that correspond to class-specific logits. In the work here, the cell collective is tasked to distinguish between objects resembling planes, chairs, cars, tables, houses, guitars, and boats. The cells are trained with cross entropy loss to predict the class label through gradient-based numerical optimization. More details on the neural network architecture and training are provided in the Materials and Methods section.


Rather than matching against a single, predefined configuration, our system generalizes across entire classes of shapes — including different e.g.\ planes, tables, etc. This shift from precise self-recognition to high-level shape classification enables greater flexibility and tolerance to variation. Crucially, the system can detect structural inconsistencies caused by missing or faulty modules, using only local interactions and without requiring actuation or centralized sensing.

We validate our approach on real hardware (Fig.~\ref{fig:overview}),  demonstrating that shape classification is not only feasible in a fully distributed setting, but also robust under the kinds of imperfections and hardware constrains that arise in the physical world. 
By enabling modular cellular bricks to autonomously infer their morphology and detect structural faults, this system brings us closer to creating artificial systems with some of the properties that helped biological organisms to strive — such as decentralized self-assessment, resilience to damage, and adaptive organization — without the need for centralized control or predefined templates. 

\section*{RESULTS}
We evaluated our approach on simple cellular  bricks, which are composed of printed circuit board (PCB) cubes with electrical connectors on their six faces, a microcontroller module, a LED to display color information (i.e.\ the class label outputted by the cell) and the electronic components necessary to power these components (Fig.~\ref{fig:overview}A).  Multiple bricks can be stacked together to create different objects (Fig.~\ref{fig:overview}B).  
 These cellular bricks can communicate and aggregate information purely locally through a custom protocol (digital serial communication). Over multiple iterations they are tasked to globally convergence on the correct shape label (Fig.~\ref{fig:overview}C). To account for unreliable communications, messages are transmitted five times between each forward pass of the network and for a period of three seconds. Invalid messages, including messages that did not arrive complete or messages without a valid header are discarded. In this way, the modules collect their neighbors information for using them as input to the NCA  (Fig.~\ref{fig:overview}D). If there is no  neighbor connected to a particular face or a valid message does not arrive in time, the input values corresponding to that face default to 0, except for the first bias element, which is always 1. Multiple cellular bricks can be stacked together to create 3D shapes.

We conducted several large-scale experiments with more than 500 bricks in simulation and almost 200 physical bricks. In simulation, the approach is able to reach an overall accuracy score of 98.97\%. Divided by object class, accuracies are 99.54\% for planes, 99.71\% for chairs, 98.6\% for cars, 99.66\% for tables, 85.04\% for houses, 96.98\% for guitars and 99.27\% for boats (Fig.~\ref{fig:convergence}A). 
To test how the approach works in the real world, we transferred the NCA trained in simulation onto the physical bricks and built four distinct shapes (Fig.~\ref{fig:overview}B) with numbers of bricks ranging from 26 for a guitar, to 197 for a round table. 

\begin{figure}[t]
  \centering
\includegraphics[width=1.0\textwidth]{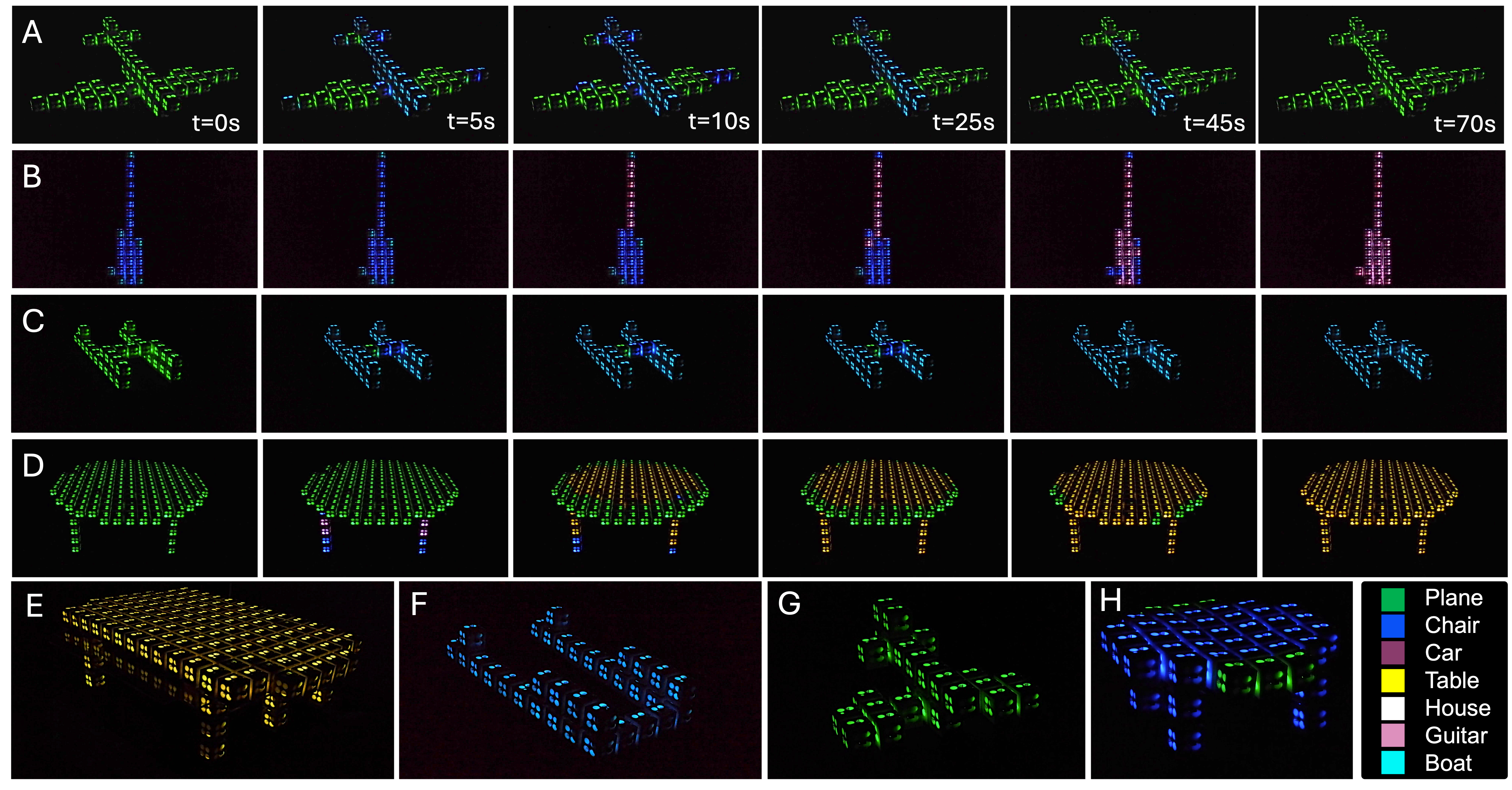}
  \caption{\textbf{Shape classification results}. \normalfont The algorithm is able to successfully classify different shapes (movie S1). Shown here are classification sequences for a plane (\textbf{A}), a guitar (\textbf{B}), a boat (\textbf{C}), and a table (\textbf{D}). In the span of 23 iterations (corresponding to 70s), all bricks  converge to the correct object class and keep the correct classification until the end of the experiment at iteration 60 (180s). Out of distribution generalization is shown in (\textbf{E}--\textbf{H}), which are shapes that the algorithm has never seen during training (movie S3). The algorithm is able to classify all of these correctly, except the smaller table (\textbf{H}) which was misclassified as a chair.  
  }
    \label{fig:shapeclassification}
\end{figure}

Fig.~\ref{fig:shapeclassification}A--D shows a timeseries of how the cellular bricks reach consensus to which object class they think they belong to, successfully recognizing a plane, a guitar, a boat, and a table; see movie S1. Self-recognition takes around three minutes in real-time. All bricks are powered up at approximately the same time when switching the power supply unit on. After checking for flags that could signal a different procedure, they start the cycle of (1) running the neural network, (2) sending/receiving their current state and their neighbors states, and repeating from (1). After 60 self-classification cycles the bricks modules stop and connect to the WiFi network to post their record of states, times and classifications for each cycle to a database. To test consistency, each classification test was run three times. 
The results show that the approach is remarkably robust, with a success rate of 100\% for the four shapes we build (i.e.\ all the cubes reach consensus on which shape they are a part of). 

\subsection*{Robustness to faulty cells}

Biological systems are remarkably robust to damage, noise, and faulty components. From cellular networks in living organisms to entire anatomical structures, biological processes often continue functioning even when faced with partial failure or incomplete information\cite{farinella1956transformation, vieira2020advancements,levin2019endogenous, vogg2019model}. This resilience stems from the local, distributed nature of biological communication and decision-making, where cells collectively assess and respond to their environment through redundant, fault-tolerant interactions.

\begin{figure}[t]
  \centering
  \includegraphics[width=1.0\textwidth]{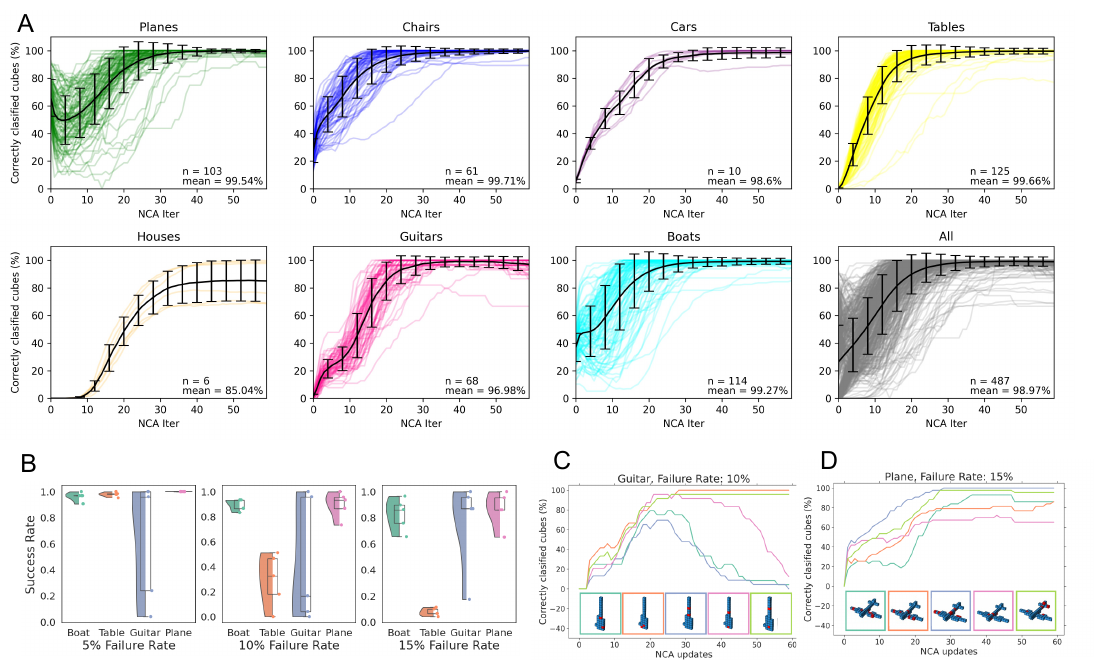}
  \caption{\textbf{Classification results in simulation and hardware}. \normalfont For all 487 shapes used for training, we show the classification convergence rate in simulation (\textbf{A}). Averaged over all shapes, we reach a classification accuracy of 98.97\%. In hardware,  we test the performance of the approach under various levels of increasing percentage of faulty cells (\textbf{B}). While convergence rate for the four shapes built is 100\%, accuracy declines under increasing levels of noise (tested in five independent runs for each shape and failure rate). Some shapes -- such as the boat and plane -- are surprisingly robust even with 15\% noise. We show examples of five guitars (\textbf{C}) and five planes (\textbf{D}) with different disabled cells (marked in red). Because of their design, the plane is much more robust than the guitar, in which a single failure along the guitar neck can disrupt the classification process.}
    \label{fig:convergence}
\end{figure}

Inspired by this property, we conducted a series of experiments to evaluate the fault tolerance of our system under simulated communication failures. Specifically, we investigated how disabling a subset of cellular bricks—preventing them from sending or receiving messages—affects shape recognition accuracy and convergence speed (Fig.~\ref{fig:convergence}B). We introduced faults to 5\%, 10\%, and 15\% of the modules randomly within each assembled shape and ran five trials per condition to assess consistency; see movie S2.

The results show that most shapes maintain high recognition performance at 5\% failure rates, suggesting that the system exhibits a level of redundancy that contributes to its robustness. 
Notably, some shapes, such as the plane and boat, showed only minimal degradation in classification accuracy even at 15\% failure rates, demonstrating strong collective robustness. However, shapes with narrow structural bottlenecks, like the guitar, were more sensitive to localized faults. In these cases, failure of a single module in the neck region could sever connectivity between subparts of the shape, leading to misclassification or delayed convergence. These findings support the hypothesis that local, learned communication rules—like those in multicellular organisms—can lead to globally coherent and robust behavior in modular cellular systems, even under imperfect conditions.

\subsection*{Robustness to out-of-distribution shape variations}

In biological systems, collective structures and functions are typically robust to variations in morphology, size, and proportion. Organs may vary in shape across individuals, limbs may regenerate at different scales, and yet the overall anatomical identity and function are preserved. This capacity for generalization—recognizing a category of form rather than a precise template — is fundamental to biological development and repair. Inspired by this principle, we evaluated whether our system could similarly generalize beyond the specific examples it encountered during training.

To assess this, we designed a series of test shapes that introduce novel variations within known shape classes; see movie S3. First, we modified a table instance from the training set by removing parts from two sides and altering the design to include five legs placed at random positions and shortened in length (Fig.~\ref{fig:shapeclassification}E). This tests the model’s ability to handle asymmetry and irregular support structures. Next, we modified a known boat configuration by shifting the central bridge structure from the middle to an off-center position, testing sensitivity to internal rearrangement (Fig.~\ref{fig:shapeclassification}F). Finally, we evaluated the system on scaled-down versions of both the plane and table configurations (Figs.~\ref{fig:shapeclassification}G, H), challenging the NCA’s capacity to infer shape class under global size reduction.

The results demonstrate that the system can successfully classify several of these novel configurations, confirming a degree of invariance to geometric variation and scale (Figs.~\ref{fig:shapeclassification}E–G). For instance, the altered table with five legs was still correctly classified as a table, and the shifted boat bridge did not significantly impair recognition. These behaviors suggest that the distributed representations learned by the NCA capture abstract structural features rather than overfitting to specific examples—mirroring biological systems' capacity to recognize morphologies despite developmental variability.

However, the system is not immune to failure. The scaled-down table, for example, was misclassified as a chair (Fig.~\ref{fig:shapeclassification}H). This may reflect a limitation in spatial resolution or context available at the smaller scale, where reduced module count compresses structural cues. Such misclassifications highlight opportunities for improving the system’s scale invariance and further aligning its behavior with biological robustness.

Overall, these experiments underscore the potential of local, learned communication rules to support generalizable morphological inference—an essential step toward building artificial systems that exhibit the flexible, fault-tolerant pattern recognition found in natural development.

\begin{figure}
  \centering
  \includegraphics[width=1.0\textwidth]{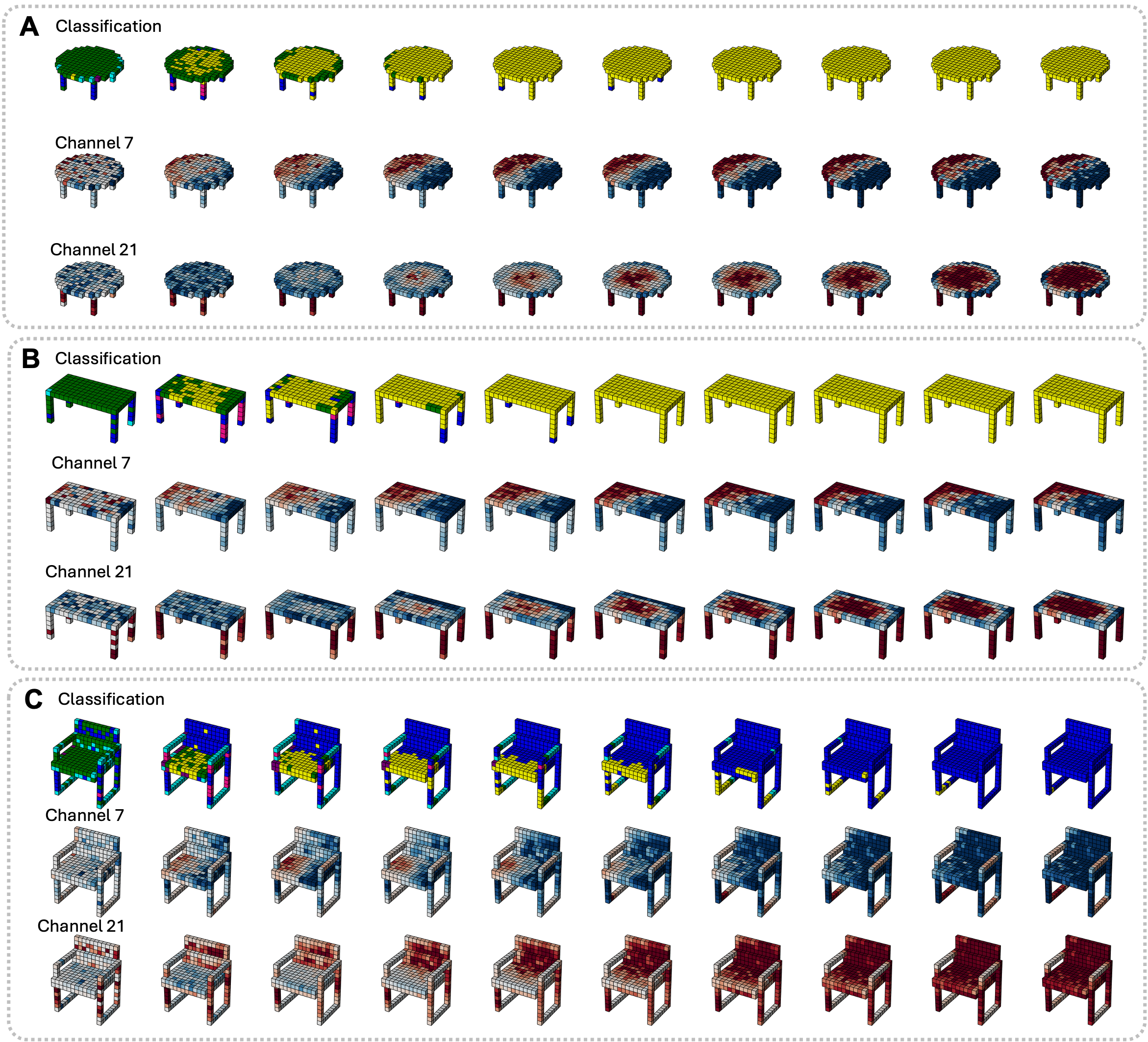}
  \caption{\textbf{Visualization of hidden channel communication}. \normalfont For three different shapes, we show the classification output and the visualization of different hidden channels over time. Tables (\textbf{A}, \textbf{B}) show a clear left-right (channel 7) and radial (channel 21)   patterning. Chairs (\textbf{C}), on the other hand, show a more pronounced anterior-posterior patterning (channel 21), which we hypothesize helps the NCA to tell apart chairs from tables. Notice how the top part of the chair in (\textbf{C})  is first classified as a table (yellow), before the cells slowly reach agreement that they are part of a chair (blue). }
    \label{fig:hidden_channels}
\end{figure}

\subsection*{Emerged communication strategies}

Collective systems can come up with efficient strategies to reach their goal. In biological organisms, this is exemplified by how cells coordinate during development to form complex structures without central control. Cells make local decisions based on their environment, often guided by morphogens—diffusible molecules that form gradients across developing tissues \cite{turing1990chemical,rogers2011morphogen,ashe2006interpretation}. These gradients provide positional information, enabling cells to infer their location within the organism and adopt appropriate identities, such as becoming part of a limb or an organ. This decentralized yet coordinated decision-making process inspired our investigation into the communication strategies developed by NCAs to recognize and differentiate shapes. Specifically, we drew inspiration from the role of morphogens in biology and examined the activation patterns of the NCA’s hidden channels.

Fig.~\ref{fig:hidden_channels} shows a sequence of visualizations of different hidden channels together with the classification output of the NCA for three different shapes (two tables and one chair) over time; see movie S4. When looking at the classification output, we can see that the cells early on reach agreement for the table designs (Fig.\ref{fig:hidden_channels}A, B). Left-right morphogen-like  (channel 7) and radial (channel 21) patternings are established early, resembling the emergence of developmental axes in embryos.

\begin{wrapfigure}{r}{0.5\textwidth}
\vspace{-0.0in}
  \centering
  \includegraphics[width=0.48\textwidth]{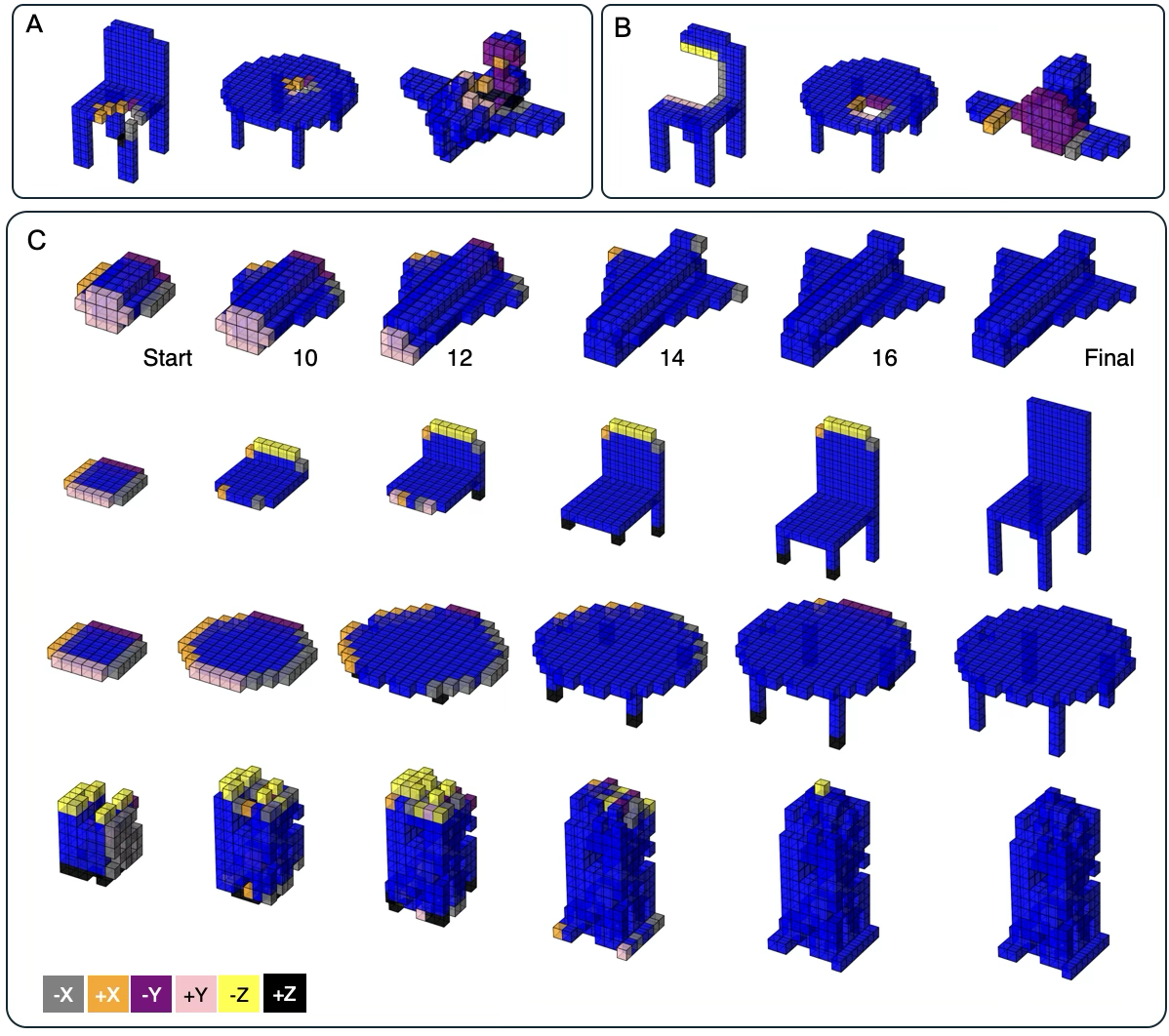}
  \caption{\textbf{Simulated damage detection and regeneration}. \normalfont
  Using spherical (\textbf{A}) and cubic damage (\textbf{B}), the model was trained to predict the direction of damage (legend at the bottom shows mapping between directions and colors). After training, the model can tell the direction of damage with high accuracy. (\textbf{C}) The same system is able to guide a shape-conditioned damage recovery process, starting from only a few seed cells, a configuration never seen during training.}
  \label{fig:damage_detection}
\end{wrapfigure}
How is the approach able to tell apart tables from chairs? Looking at the morphogen development for a chair object can give some insights (Fig.~\ref{fig:hidden_channels}C). Similar to the tables, many cells in the chair are initially classified as a plane (green). However, unlike in the table case, an anterior-posterior patterning is established (channel 21), akin to the biological head-to-tail axis. This pattern is mirrored in the classification of the cells, which initially identify the backrest and seat as separate table-like components. Over time, the signal propagates from posterior to anterior, guiding the cells to reach consensus that they form a chair. These observations suggest that the default classification for tables and chairs is initially “table,” but morphogen-like signals originating from the backrest region of chairs gradually induce a reclassification in neighboring cells, leading to a coherent identification of the object as a chair.

\subsection*{Damage detection and recovery}
An exciting possibility is to not   only classify 
which shapes  the  cell is a part of but also for each cell to determine if any damage  occurred to its neighbors.  
To eliminate ambiguity (e.g.\ a particular damaged table might look like an undamaged chair), here we conditioned the NCA on a particular shape instead of a general class. 
To do this, we simply replaced the class embeddings with a compact 3D convolutional encoder that takes in a voxelized shape and returns an embedding. The convolutional encoder consisted of a convolutional layer that reduced the voxel shape from 15$\times$15$\times$15 to 5$\times$5$\times$5, then flattened it, and projected the reduced shape into the cell hidden dimension size. The number of parameters varied, but the maximum size of the cell hidden dimensions (128) was around 10,000 parameters. 
Given this conditioning, each cell was trained to predict a single damage direction through cross-entropy loss. Specifically, each cell was classified as either not having any damaged neighbors or as having damage in the $-X$, $+X$, $-Y$, $+Y$, $-Z$, or $+Z$ direction. 

 We trained on synthetically altered shapes, with both spherical  (Fig~\ref{fig:damage_detection}A) and cubic damage (Fig~\ref{fig:damage_detection}B). For each training iteration, we sampled random shapes, damaged them with either one of the two types of damage, and ran the NCA for a random number of steps (between 64 and 96). The last channels for each cell was treated as logits for classification, and the model was optimized  using the labels generated by the synthetically damaged shapes.  
 We achieved a damage detection accuracy of more than 90\% for all the different types of shapes it was tested on.

Can we exploit this ability to also recover from damage, instead of merely predicting it? In this regard, biological systems possess remarkable capabilities, often through distributed, decentralized processes of sensing and regeneration. For example, organisms like planarians and axolotls exhibit the ability to regrow complex body parts by coordinating cell behavior in response to injury, even in the absence of a central controller \cite{tanaka2011cellular,reddien2004fundamentals}. These processes involve local detection of damage, activation of repair pathways, and iterative remodeling until the original structure is restored. 

\begin{wraptable}{r}{0.4\textwidth}
  \vspace{-\baselineskip} 
  \caption{Recovery accuracy for different hidden dimension sizes and different object classes.}
  \label{tab:recovery_accuracy}
  \centering
  \begin{tabular}{lcccc}
  \toprule
  \textbf{Class / \#Hidden} & \textbf{20} & \textbf{48} & \textbf{96} & \textbf{128} \\
  \midrule
  Aircraft & 0.48 & 0.63 & 0.89 & 0.91 \\
  Boat     & 0.68 & 0.82 & 0.93 & 0.95 \\
  Car      & 0.63 & 0.89 & 0.88 & 0.93 \\
  Guitar   & 0.23 & 0.41 & 0.55 & 0.67 \\
  House    & 0.21 & 0.37 & 0.51 & 0.64 \\
  Table    & 0.57 & 0.66 & 0.81 & 0.92 \\
  Chair    & 0.53 & 0.67 & 0.84 & 0.93 \\
  \bottomrule
  \end{tabular}
\end{wraptable}
To test the ability of our system to guide damage recovery, we started from a small cluster of cells, and added cells in the direction determined by the existing cells; this process was repeated  until no more damaged was  detected. Finally, we determined  the recovery accuracy as the difference between the final regrown shape and the original target shape.

Surprisingly, without ever being trained with only a few cells, the model was able to detect damage and recover almost all shapes across all object classes with a high accuracy; a selection of shape recovery sequences are shown in  Fig~\ref{fig:damage_detection}C.  We  compared the approach across different cell hidden dimension sizes (Table~\ref{tab:recovery_accuracy}) and found that models with larger hidden states performed  significantly better. 
These results are likely explained by the fact that the larger hidden states allow the models to capture more information across development. While recovery accuracy rates were  generally high, the guitar and the house posed to be the most challenging object classes (e.g.\ an example recovery sequence for a particular house is shown in the bottom row in Fig~\ref{fig:damage_detection}C). 

\section*{DISCUSSION}

The experiments presented here demonstrate that learned, fully-decentralised control enables large assemblies of simple, identical modules to reach a coherent understanding of their global morphology and to pinpoint structural faults, using nothing more than short-range communication and on-board computation. In hardware trials with almost 200 cellular bricks, the collective converged on the correct class for 3D shapes such as a plane, table and car in fewer than 60 update cycles—about three minutes of real time. Extending the same NCA to localize missing modules shows that damage detection and even damage recovery can be achieved with exactly the same infrastructure and identical code on every cube. Taken together, these results constitute the first physical realisation of large-scale, bio-inspired self-recognition in three dimensions. By allowing decentralized self-recognition, the approach thus takes a step towards the major goal of developing modular self-reconfigurable  systems\cite{yim2007modular}. 

\subsection*{Relation to biological and robotic systems}

Unlike previous modular-robot platforms that depend on centralised computation or handcrafted rules, our approach mirrors the decentralised strategies seen in morphogenetic processes—from planarian regeneration to salamander limb regrowth—where global form emerges from purely local interactions. Kilobots and similar swarms \cite{rubenstein2014programmable} have illustrated consensus in two dimensions, but they require line-of-sight signalling and prescriptive behaviour tables; their extension to 3D structures has proved challenging. By coupling 3D NCAs with minimal six-face connectivity, we close this gap and show that a biologically plausible mechanism scales to hardware assembled from off-the-shelf microcontrollers and passive connectors. Moreover, the collective generalises beyond the discrete instances encountered during learning: shortened table legs and rescaled vehicles were still classified correctly, highlighting an ability to infer shape categories rather than match one-to-one templates. Such categorical reasoning is fundamental to developmental robustness in living tissues, and its replication in machines opens a path toward morphologically resilient artefacts.

\subsection*{Robustness and emergent communication}

The network tolerates realistic imperfections. For some shapes, classification convergence is still high, even with up to 15\% of modules rendered silent—either through message loss or complete failure. Inspired by morphogen-driven development in organisms, we observed that hidden channels within the NCA form spatial gradients reminiscent of biological axes, enabling cells to coordinate classification decisions without centralized control. The emergence of consistent left-right and radial patterns in tables, and an anterior-posterior pattern in chairs, suggests that the NCA leverages internal morphogen-like signals to disambiguate similar shapes. Notably, the gradual reclassification of chair components from "table" to "chair" highlights a dynamic, context-sensitive communication process, mirroring how biological cells interpret positional cues. These results support the idea that biologically inspired mechanisms can guide the design of robust and interpretable collective systems.

\subsection*{Limitations}

Incorporating actuation — e.g.\ magnetically docked milli-scale blocks or lattice-based walkers — will be required for active self-repair in the future. Additionally, miniaturisation of the electronics and refinement of the mechanical connectors \cite{ze2022spinning, zhakypov2019designing} would permit denser collectives and more organic geometries, bringing the platform closer to the cellular scale of its biological inspiration. Closed-loop growth, whereby bricks autonomously recruit additional bricks from a pool, could transform the current recognition capability into full morphological regulation. 

\newpage
\section*{Supporting Information}
\renewcommand{\thefigure}{S\arabic{figure}}
\setcounter{figure}{0}

This sections includes:
\begin{itemize}
    \item Supporting text
    \item Figs. S1 -- S6
    \item Legends for movies S1 -- S4
\end{itemize}

Other supporting materials for this manuscript include the following:
\begin{itemize}
    \item Movies S1--S4
\end{itemize}

\section*{Supporting Information Text}
Here we detail the  model architecture, training process, and hardware system we developed.  We also describe the mechanical assembly, electronics, and communication protocol used to construct and operate these modular devices. We will release all the hardware specification and source code for easy reproducibility of our results.

\subsection*{Training Data}
\begin{figure}[H]
  \centering
  \includegraphics[width=1.0\textwidth, trim=200 105 150 90, clip]{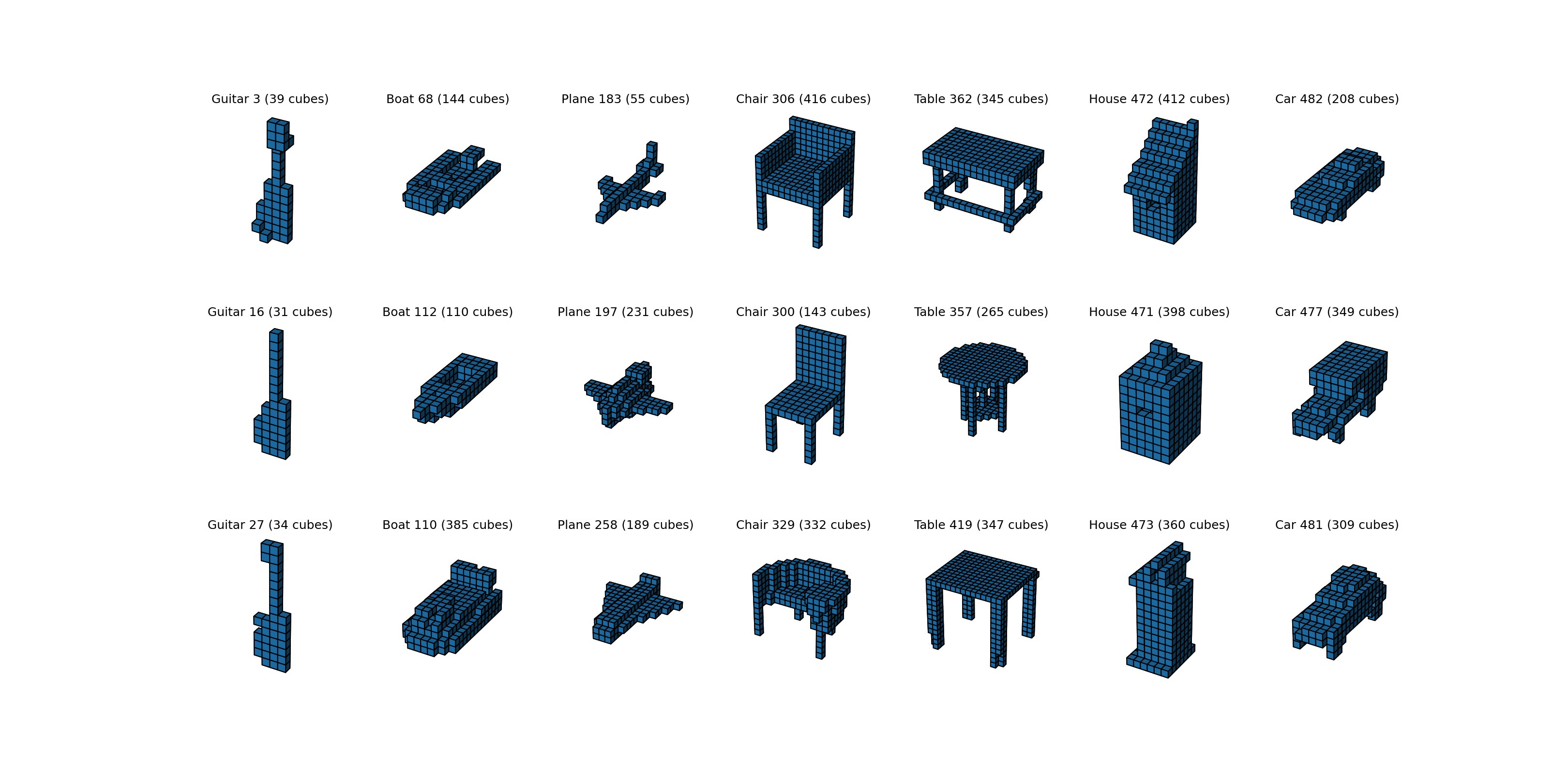}
  \caption{\textbf{Examples of shapes included in the training set}. \normalfont The training set contains 487 instances of 7 different shapes (68 guitars, 114 boats, 6 houses, 125 tables, 10 cars, 61 chairs and 103 planes). The size of the instances varies from 16 to 562 cubes.} 
    \label{fig:training_set}
\end{figure}

The training dataset comprised voxelized binary representations of 3D shapes, each associated with one of seven discrete class labels. In more detail, we use a subset of the ShapeNet dataset \cite{chang2015shapenet}, including the following object classes: aircraft, boat, car, guitar, house, table and chair. Because of the number of available hardware cubes, we reduce the resolution of the objects to a maximum of 15$\times$15$\times$15 cells. Afterwards, we manually filter designs that are not continuous anymore or that got unrecognizable at the lower resolution. Some examples are shown in Fig.~\ref{fig:training_set}. 

\subsection*{Neural Network Architecture}

The model in this work is a three-dimensional Neural Cellular Automata (3D NCA) network designed to perform volumetric shape classification through the iterative evolution of a spatially distributed cell state. Each voxel in the grid holds a 28-dimensional state vector, composed of an alpha channel that indicates if the cell is ``dead'' or ''alive'', twenty hidden channels that serve as internal memory, and seven channels corresponding to class-specific logits. The network consists of two main computational modules: a perception mechanism and a state update mechanism.

During the perception phase, each cell gathers information from its spatial neighborhood using a three-dimensional convolutional layer with a \(3 \times 3 \times 3\) kernel and ReLU activation. This layer outputs 84 channels (three times the number of input channels) and uses a fixed binary kernel mask to restrict the receptive field to a biologically inspired cross-shaped local pattern, emphasizing the cell itself and its immediate neighbors.

The resulting feature map is passed through a state update module composed of two consecutive \(1 \times 1 \times 1\) convolutional layers. The first of these uses ReLU activation and maintains the dimensionality at 84 channels. The second is initialized with zero weights and projects the features down to 27 channels, corresponding to all components of the cell state except the alpha channel (indicating if a cell is ``dead'' or ''alive''). The update is added to the current cell state only for cells that satisfy two conditions: their alpha channel must exceed a predefined living threshold of 0.1, and in simulation a randomly sampled condition must fall below the firing rate of 0.5. These stochastic updates introduce robustness and encourage emergent coordination across the cellular system. To ensure stability, the state update is bounded via a hyperbolic tangent activation. In physical tests, the firing rate is set to one to guarantee deterministic results and to speed up convergence.

After a number of evolution steps, the last seven channels of each cell's state vector are interpreted as class logits. These are used for per-voxel classification, but the loss and accuracy are computed only over active voxels---those that belonged to the original input shape---thereby excluding background regions from supervision and evaluation.

\subsection*{Neural Network Training}

A 3D Neural Cellular Automata (NCA) model was trained for the task of volumetric shape classification. The model operates by iteratively updating a distributed cellular state over time. Each shape was represented as a binary 3D grid indicating the presence of structure within a fixed volume. To enable voxel-wise supervision, class labels were spatially broadcasted over the active voxels of each shape. This approach encouraged the model to focus learning on regions of the grid corresponding to meaningful shape structure.

At each training step, the model was initialized with a batch of shapes and evolved over a randomly selected number of update iterations (between 60 and 120). After evolution, the final state was used to produce a voxel-wise class prediction, from which a classification decision was inferred.

The model was trained using a sparse categorical cross-entropy loss applied only to voxels marked as active in the original shape, effectively masking inactive regions from the loss computation. This loss encouraged accurate class assignment specifically within the occupied regions of the shape volume. Gradients were computed using automatic differentiation, clipped to prevent exploding gradients, and applied using the Adam optimizer with a fixed learning rate of $10^{-4}$.

Training proceeded over 4000 iterations with a total of 487 shapes, distributed in 68 guitars, 114 boats, 6 houses, 125 tables, 10 cars, 61 chairs and 103 planes. In each iteration, the dataset was randomly shuffled and partitioned into mini-batches. For each batch, the model was evolved, predictions were computed, and the loss and voxel-wise classification accuracy were measured. Accuracy was defined as the proportion of correctly classified active voxels. 

For damage detection, we train the model for 4,000 epochs at batch size 12 using  synthetically damaged shapes, incorporating spherical and cubic damage types.  In each training iteration, we randomly select shapes, apply damage, and run the NCA for a randomly chosen number of steps between 64 and 96. The final channels of each cell are interpreted as classification logits, and the model is optimized using damage direction labels derived from the corresponding damaged shapes.

\subsection*{Mechanics}

\begin{figure}[t]
  \centering
  \includegraphics[width=0.5\textwidth]{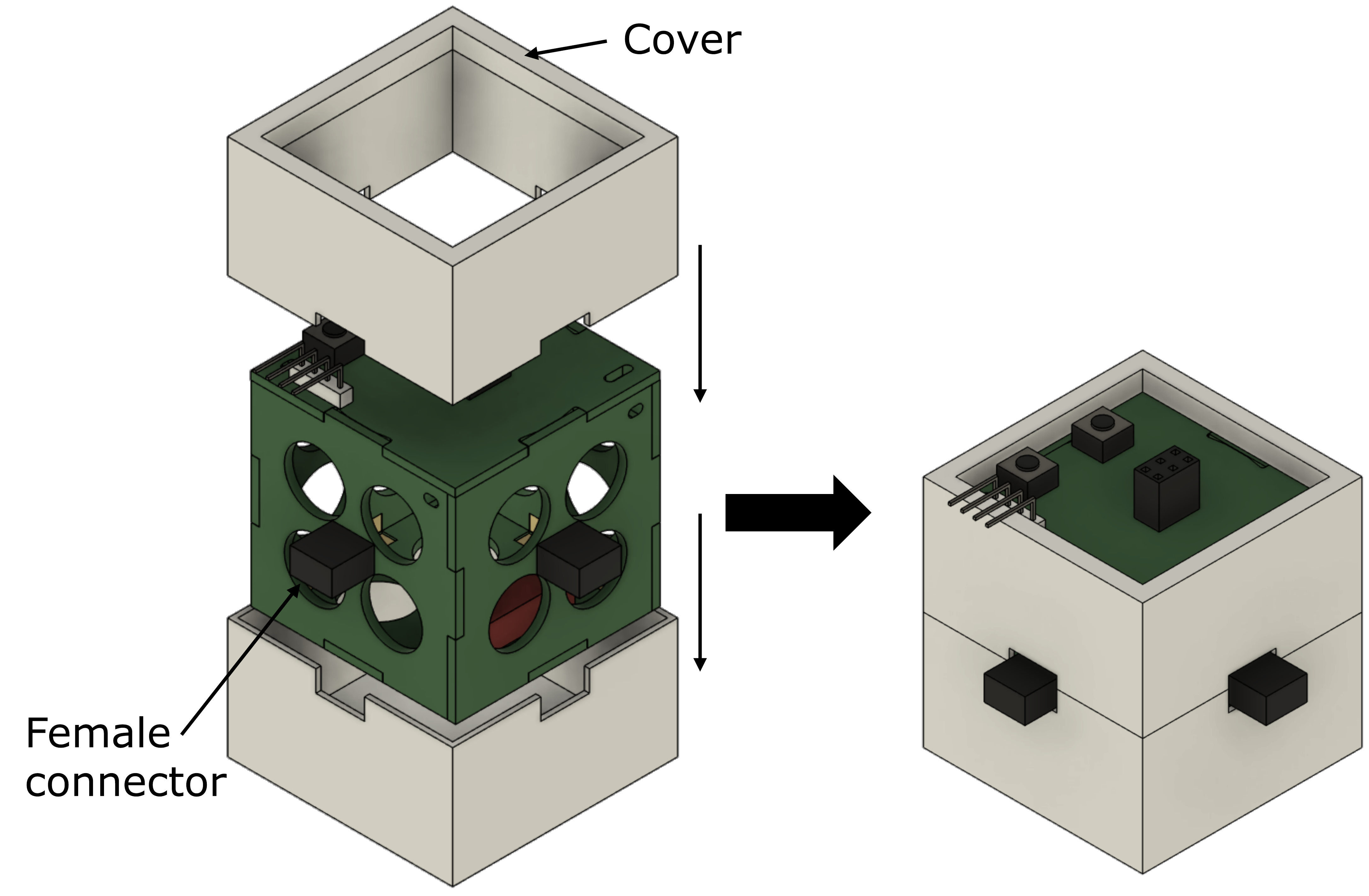}
  \\
  \vspace{0.2in}
  \includegraphics[width=0.5\textwidth]{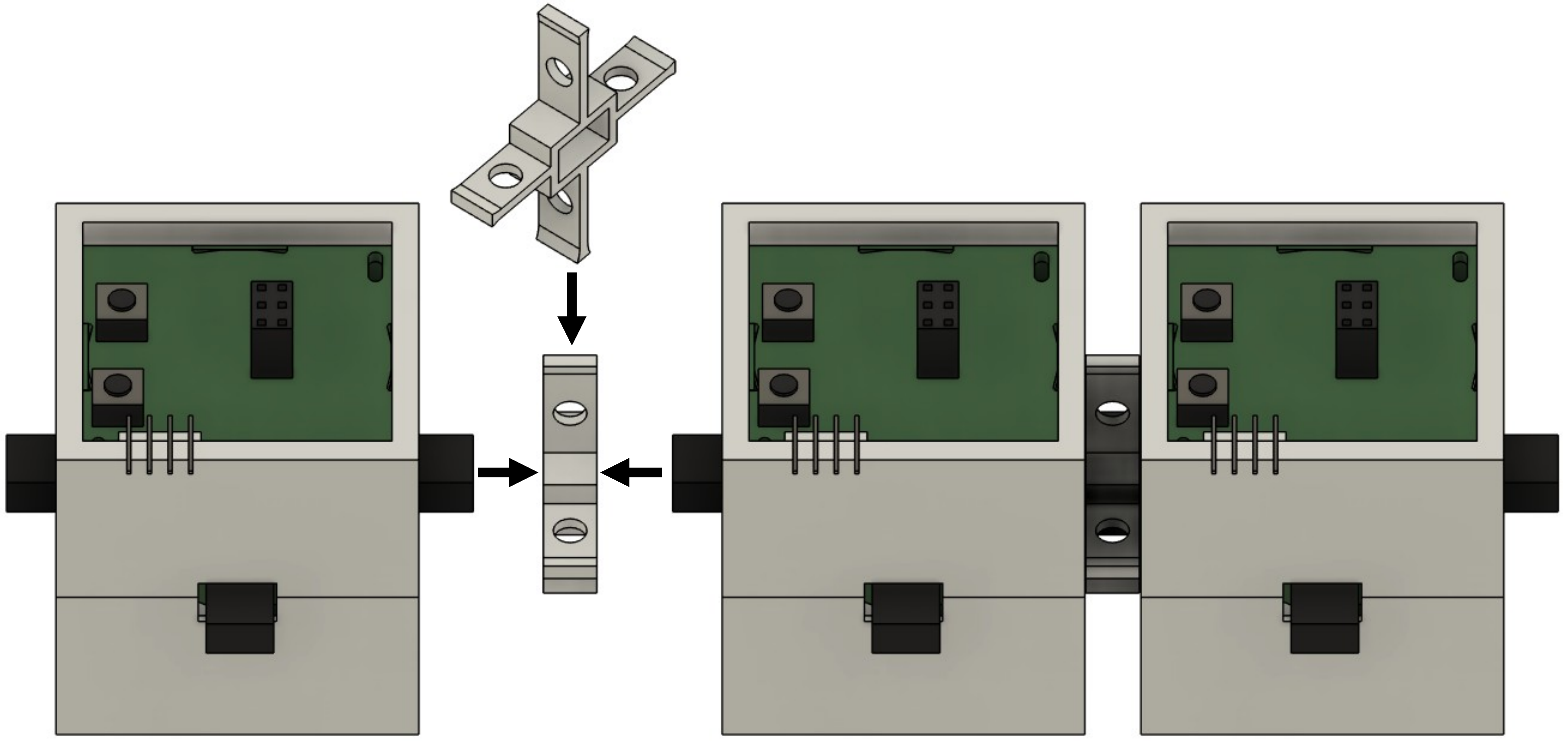}
  \caption{\textbf{Mechanics}. \normalfont (Top) The five electronics boards of the bottom of the bricks are soldered together and the top board carrying the MCU is assembled with cable ties. After that, a 3D printed white cover is assembled to diffuse the light. (Bottom) In large assemblies, a spacer is used in between bricks to distribute the load more evenly among the cube face.}
    \label{fig:mechanics}
\end{figure}

For an overview of the mechanics design see Fig.~\ref{fig:mechanics}. Using a technique for using PCBs as structural elements similar to the one in\cite{moreno2021Emerge}, one cube is assembled using 6 FR4 1.6 mm printed circuit boards (PCBs) as sides. The edges of the boards are slotted so they can fit together. Five of the boards are passive and contain only connectors. These 5 boards are soldered together by using plated pads at each board edge, which provides structural rigidity and forms the bottom of the cube. The sixth board, carrying the electronics, is placed on top and fixed to the bottom of the cube using cable ties. Each face has a doubled row 6 pin 2.54 mm header or socket connector that mates with the corresponding connector in the neighboring cube. These connectors are the only mechanical link between cubes. Two covers are printed in white PLA (polylactic acid). After they have been assembled, the edges of the covers are melted with a soldering iron tip to fix them in place. The walls of the cover are very thin and diffuse the light from the LED inside the cube. For larger cube assemblies, an extra plastic spacer is placed in between cubes to spread the load more evenly among the cube face and cover. For big shapes with a large overhanging and a single layer of modules (such as the table in Figs.~\ref{fig:convergence}D, E), a base plate was used to increase the stability.

\subsection*{Electronics}

\begin{figure}[t]
  \centering
  \includegraphics[width=0.7\textwidth]{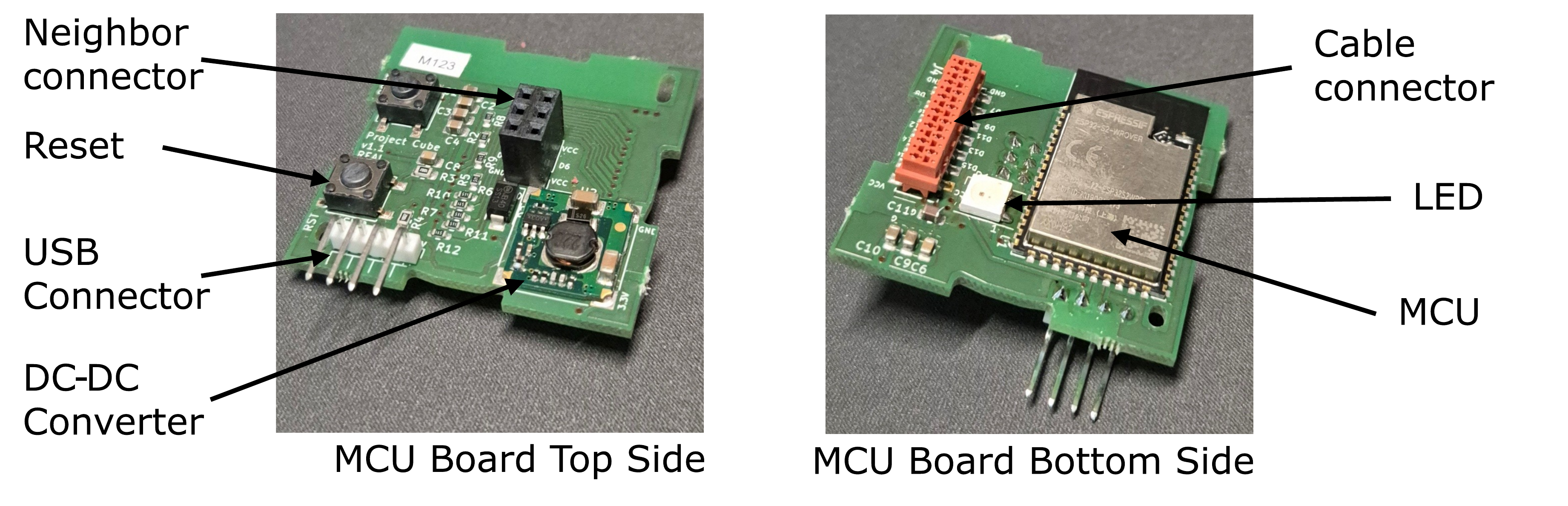}
  \\
  \vspace{0.1in} 
  \includegraphics[width=0.5\textwidth]{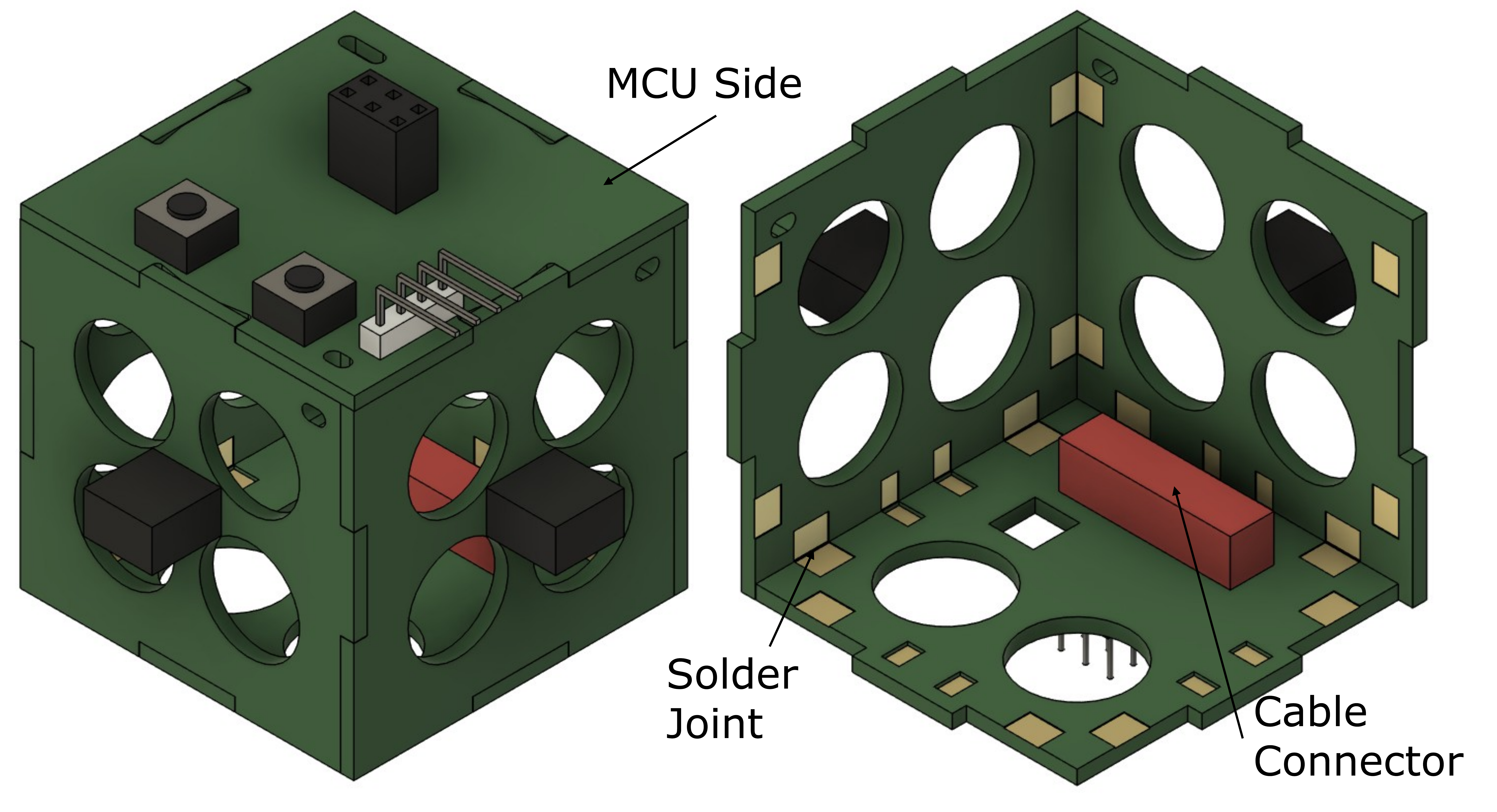}
  \caption{\textbf{Electronics of the cellular bricks}. \normalfont (Top Right) The bottom side of the microcontroller (MCU) board contains an ESP32-S2 WROVER Wifi module that manages all the processing and communication in the cube modules. Next to it lies a WS2812B RGB LED and a 14 way flat cable connector. (Top Left) The top side contains a diode protected DC-DC converter that powers the ESP32 module and the LED, and the outside connectors to USB and the cube neighbors, as well as the reset button and bootloader mode button. (Bottom) Four FR4 boards containing a header or socket connector are used as lateral sides. A fifth board containing also a 14 way micro-match connector fits the bottom. Plated pads on the edges are soldered together to provide structural rigidity and power and communication connections.}
    \label{fig:electronics}
\end{figure}

A detailed view of the electronics of the cellular bricks are shown in Fig.~\ref{fig:electronics}. All boards contain electronic traces that allow power and communications to get through to the neighboring modules. Two pins in the header or socket connector in each face are used for connecting the power supply voltage at up to 36 V, two are for ground and the remaining two are used for serial communication in a duplex (transmit-receive) way. The passive board at the bottom of the cube contains a 14-way TE micro-match ribbon cable connector to distribute power and communications between the top board, which contains the MCU, and passive boards at the bottom of the cube. The power and communications between the passive board at the bottom and the passive boards on the sides are connected with solder joints. The passive boards and cable assembly are designed to carry up to 3 A of current to all neighbors. 

The active board at the top of the cube contains a 32 bit single core microcontroller module (ESP32-S2 WROVER, Espressif Systems, China) with 4MB of flash memory, 2 MB PSRAM and 802.11 b/g/n wifi communication. The board also has an  programable RGB LED (WS2812B, WorldSemi, China) to display color information and another 14-way TE micro-match ribbon cable connector. To decrease the current through the connectors of the cubes, a DC-DC converter (K7803JT-500R3-SMD, Mornsun, China) is used. The input voltage can range from 12 to 45 volts to produce a 3.3 V output. A diode (SS13, Onsemi, USA) is also placed at the input of the DC-DC converter to prevent inverse voltage connections. The programming of the ESP32 module is done in parallel over the air (OTA) using the Wi-Fi connection, which speeds up the programming process when multiple cubes need to be reprogrammed. 

The assemblies are powered externally by connecting a power supply unit at 30 V to one or two of the cubes in separate locations around the structure (depending on the number of cubes; big shapes like the table need two power connections to not exceed the maximum current in the ribbon cable and cube connectors). 

Communication between bricks is performed by using a custom protocol digital serial communication. This protocol was used instead a standard serial (such as UART) in order to reduce the cost and size of the microcontroller as we need 6 serial buses (one per face). Our custom protocol employs the Pulse Width Modulation (PWM) hardware of the microcontroller (LEDC peripheral) to generate precise pulses and change interruptions to count the time of these pulses. 0 and 1 states are differentiated by the length of their pulses in microseconds. For either a 0 or 1 to be transmitted the line changes from a low (0 V) state to a high (3.3 V) state during a predetermined amount of time. Ones are two times longer than zeroes. A third state which is three times longer than a zero is used as a header to validate the message start. Each  number is transmitted as 32 bit floats. To account for unreliable communications, messages are transmitted five times between each forward pass of the network and during a maximum of 3 seconds. Invalid messages, including messages that did not arrive complete or messages without a valid header are discarded. 

\subsection*{Communication test and data collection}
To ensure that all data connections between bricks are working correctly we prompt all bricks to send their unique identifier number through all their faces. To achieve this we implemented the sending of a special message in which all the bytes represents the same number when we press the boot mode button on the MCU board. The message is sent to the neighboring modules which in turn change their mode and propagate the message. In this way, we can prompt all modules to send their identifier by pressing a single button, without regard to the number of modules.

We use the MAC number from the WiFi module as unique identifier. The received information is collected in a computer by requesting it from an http server  running in the microcontroller (using the ESPAsyncWebServer Arduino library), and a python script analyzes the connections and compares it to the intended shape, detecting missing connections and reconstructing the shape from the existing ones. This same server is used to toggle the different functional modes, from running the classification neural network to getting into a \emph{failure} state. The mode information is stored in non volatile memory to ensure that all modules start in the correct mode after a reset. The current mode can only be changed while the cubes are connected to the Wifi network, which only happens after a classification test or when prompting the cubes to send their id. The Wifi peripheral of the microcontroller is deactivated during tests to prevent bad interrupt timings that could affect communication.    
During a classification test the internal cell state, the current guess, the current time in ms from when the bricks started and whether the cube is in a failure state are stored at each forward pass step. After the 60 steps the cube connects to a central server in the WiFi network using an http client (using the HTTPClient Arduino library) and posts all this information to a database.

\subsection*{Detailed results on robustness to faulty cells}

\begin{figure}[H]
  \centering
  \includegraphics[width=0.3\textwidth]{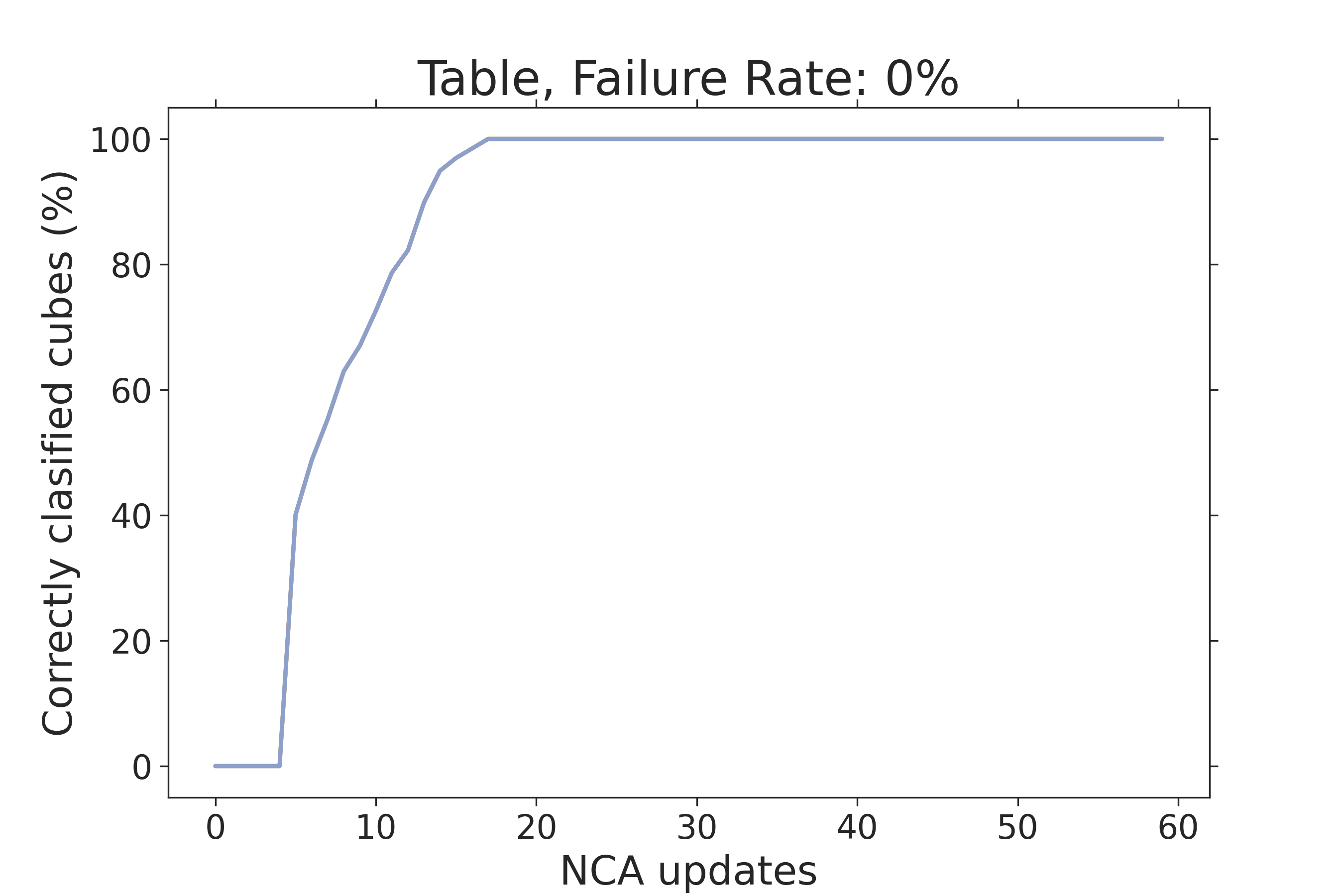}
  \includegraphics[width=0.3\textwidth]{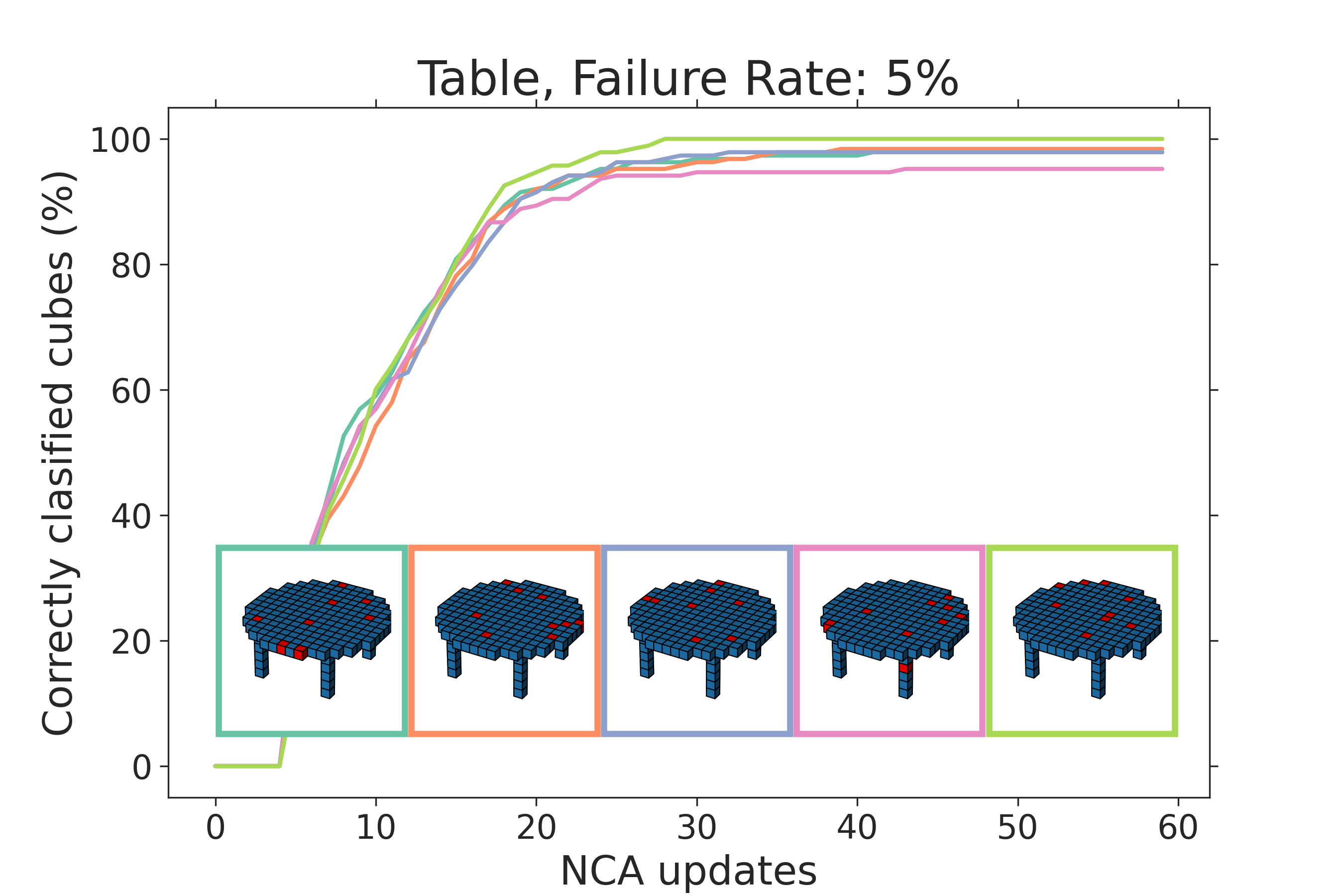}
  \includegraphics[width=0.3\textwidth]{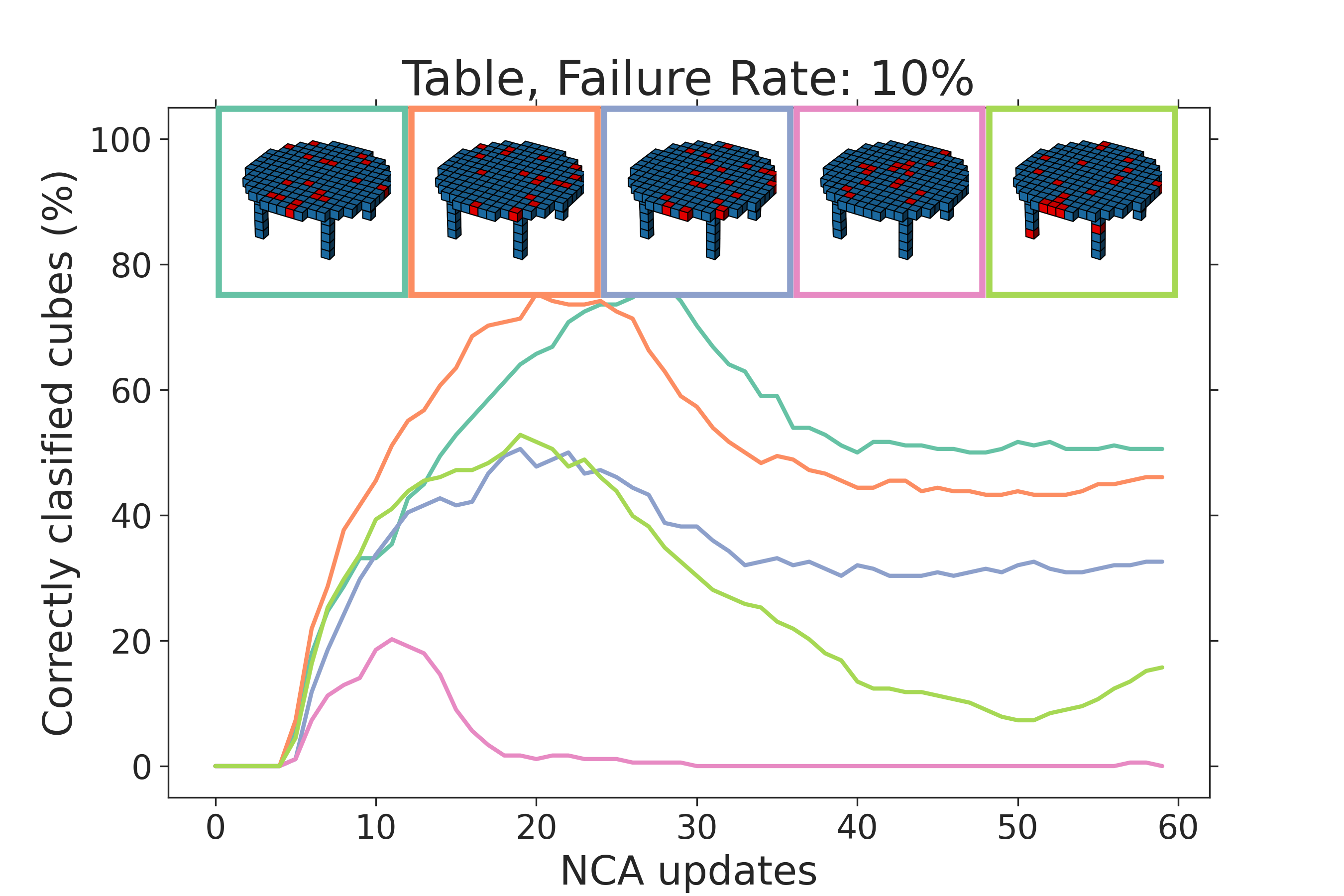}
  \includegraphics[width=0.3\textwidth]{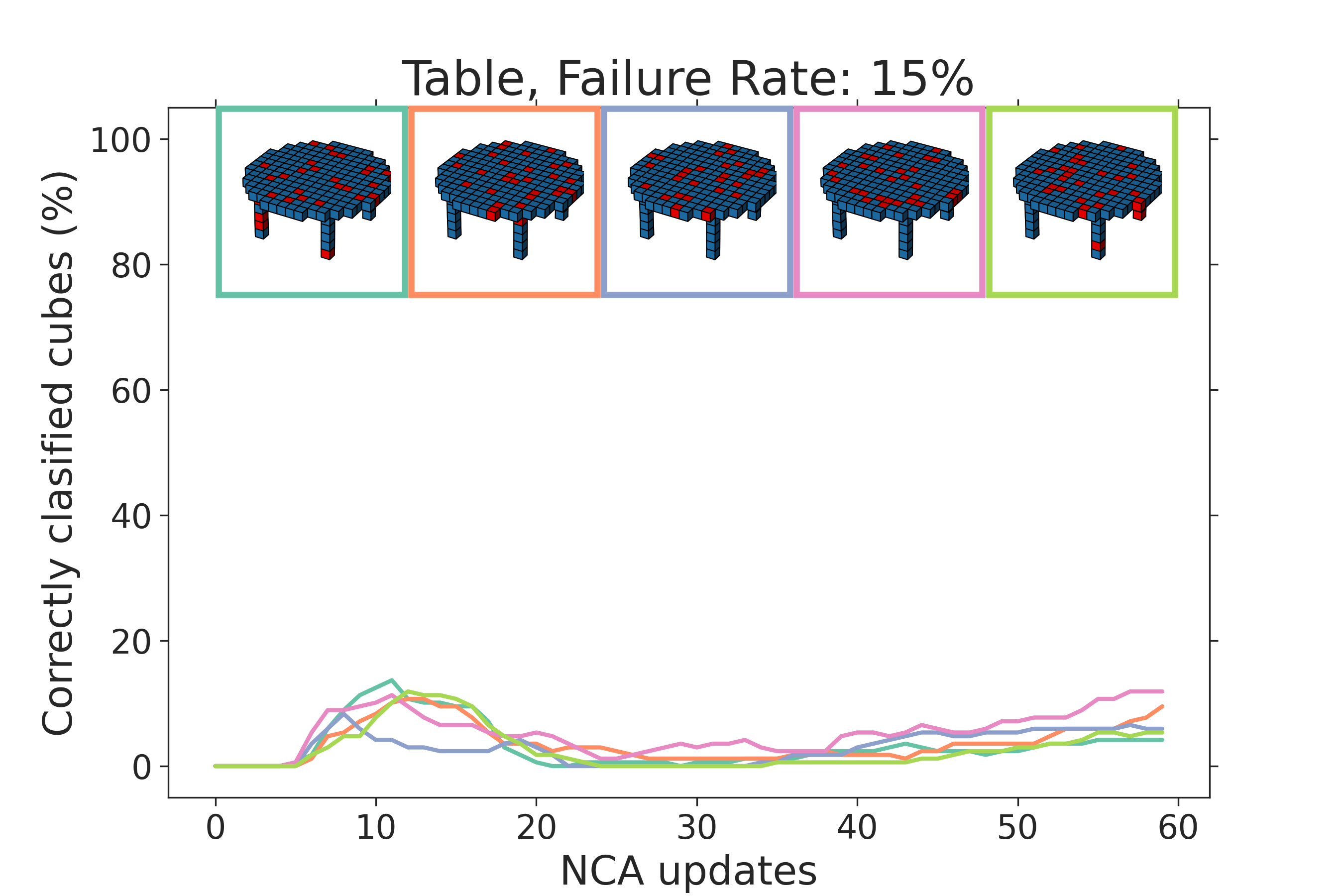}
  \caption{\textbf{Classification experiments for the table in hardware}. \normalfont The four graphs show the percentage of correctly classified cubes for failure rates of 0, 5, 10 and 15\%. The damaged modules for each experiment are indicated with small icons inside the graphs.} 
    \label{fig:table_failure_rate_table}
\end{figure}

\begin{figure}[H]
  \centering
  \includegraphics[width=0.3\textwidth]{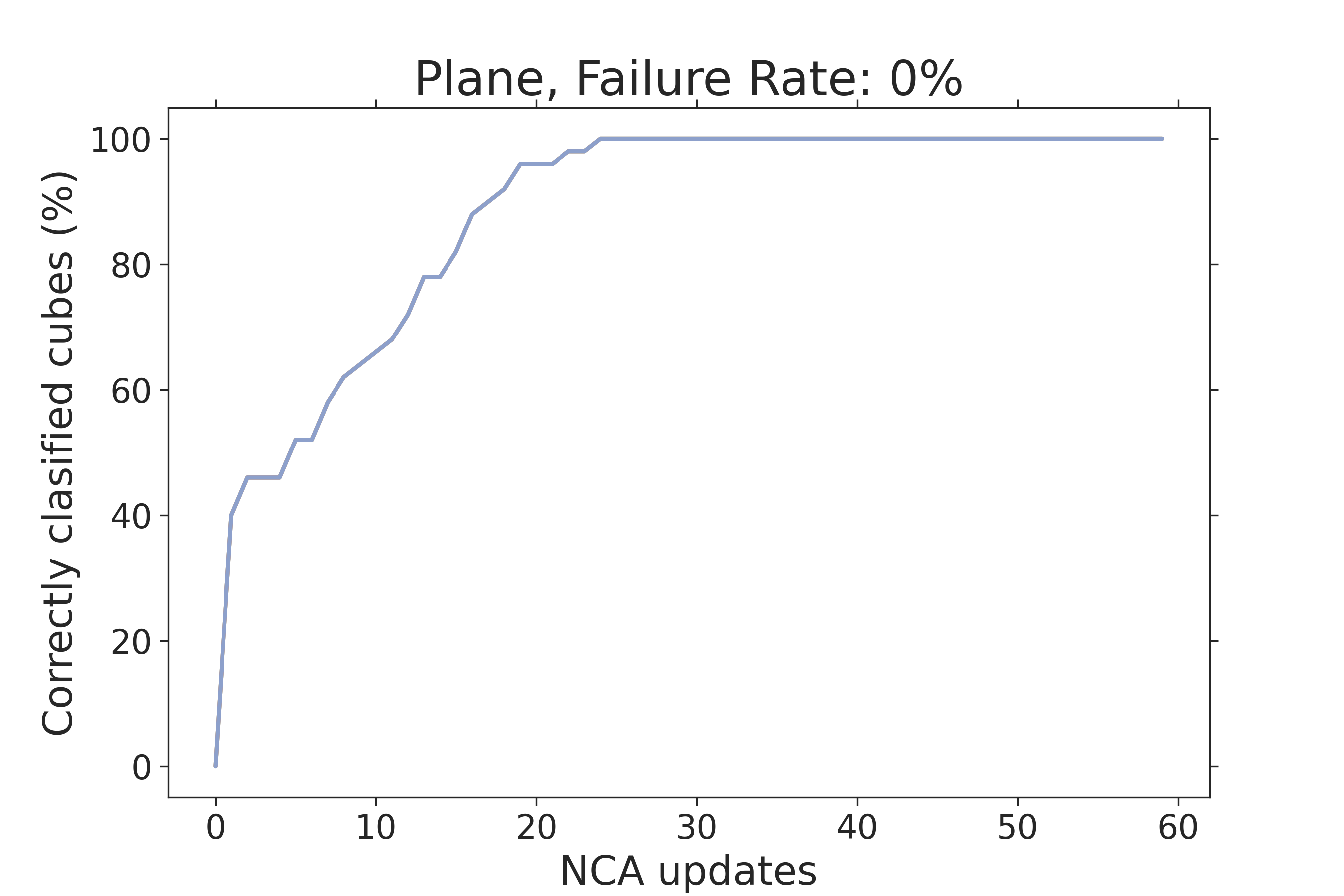}
  \includegraphics[width=0.3\textwidth]{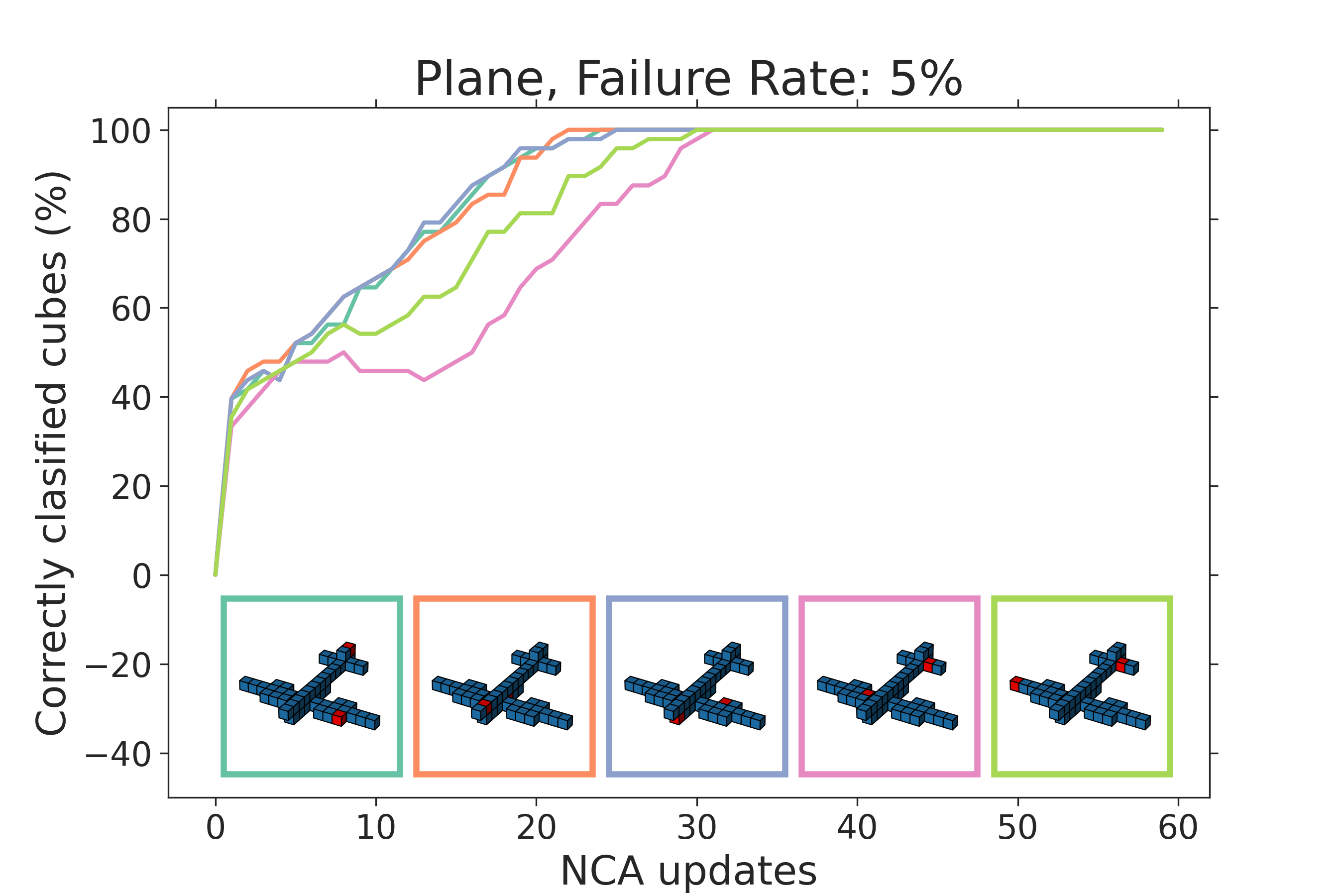}
  \includegraphics[width=0.3\textwidth]{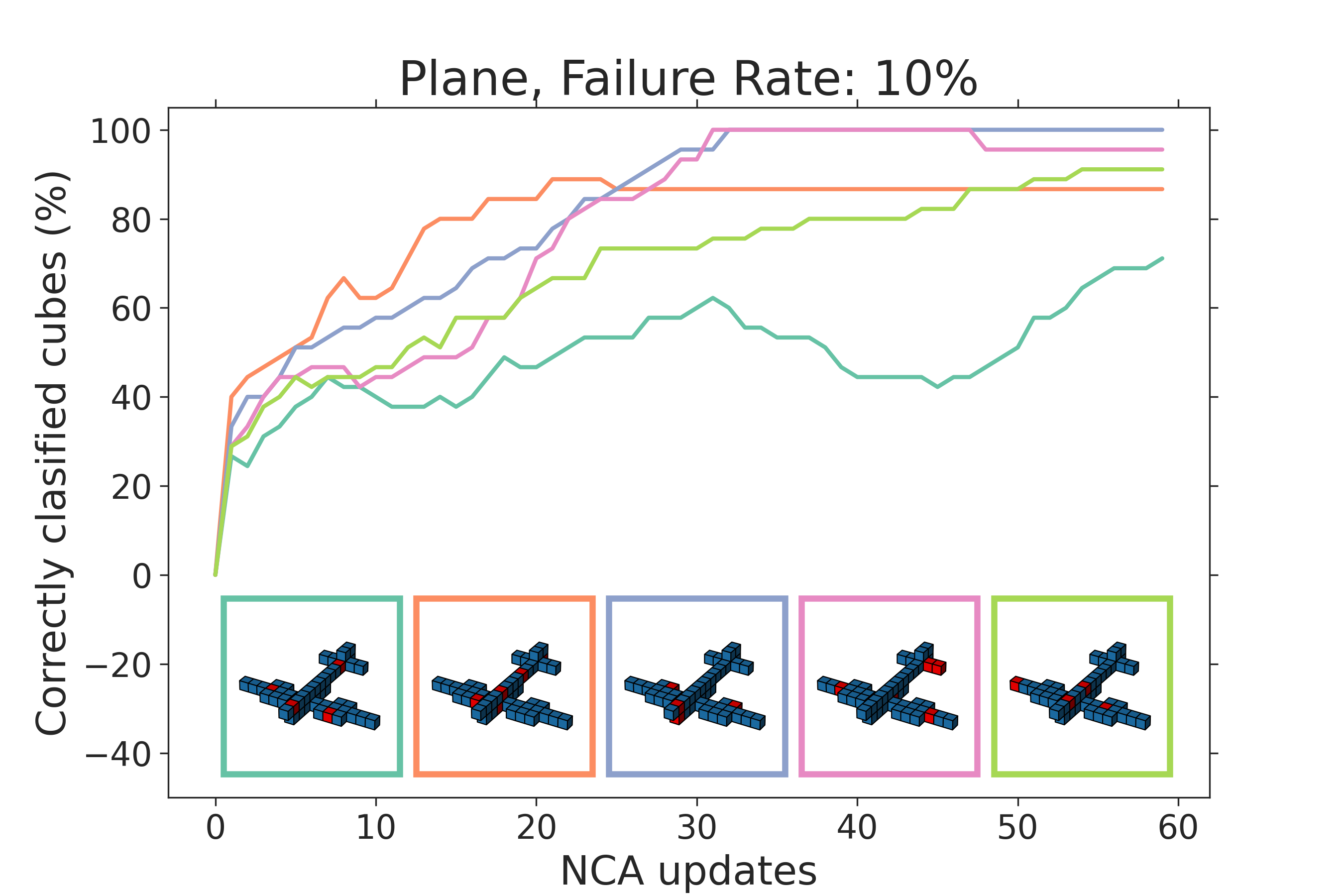}
  \includegraphics[width=0.3\textwidth]{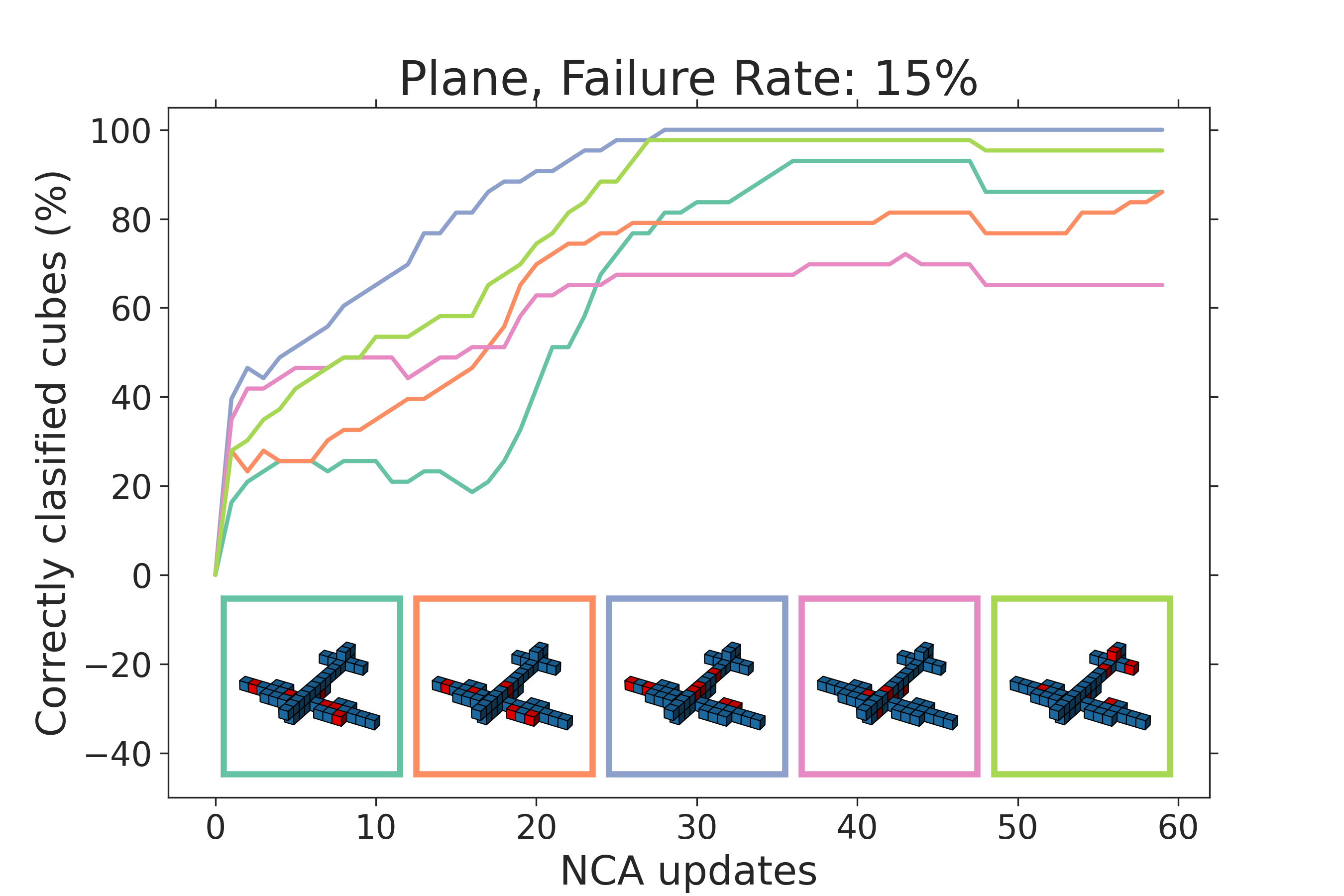}
  \caption{\textbf{Classification experiments for the plane in hardware}. \normalfont The four graphs show the percentage of correctly classified cubes for failure rates of 0, 5, 10 and 15\%. The damaged modules for each experiment are indicated with small icons inside the graphs.} 
    \label{fig:table_failure_rate_plane}
\end{figure}

\begin{figure}[H]
  \centering
  \includegraphics[width=0.3\textwidth]{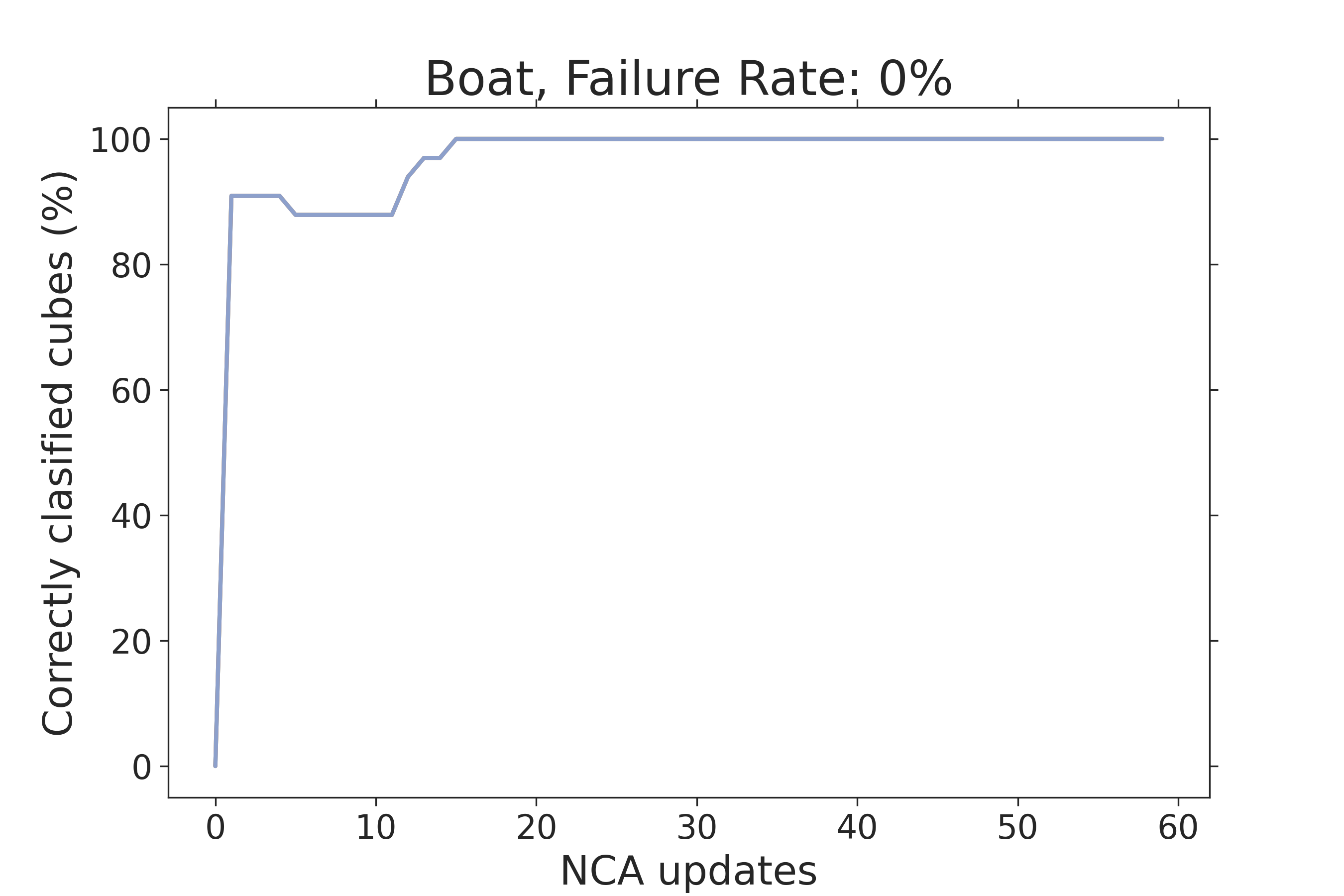}
  \includegraphics[width=0.3\textwidth]{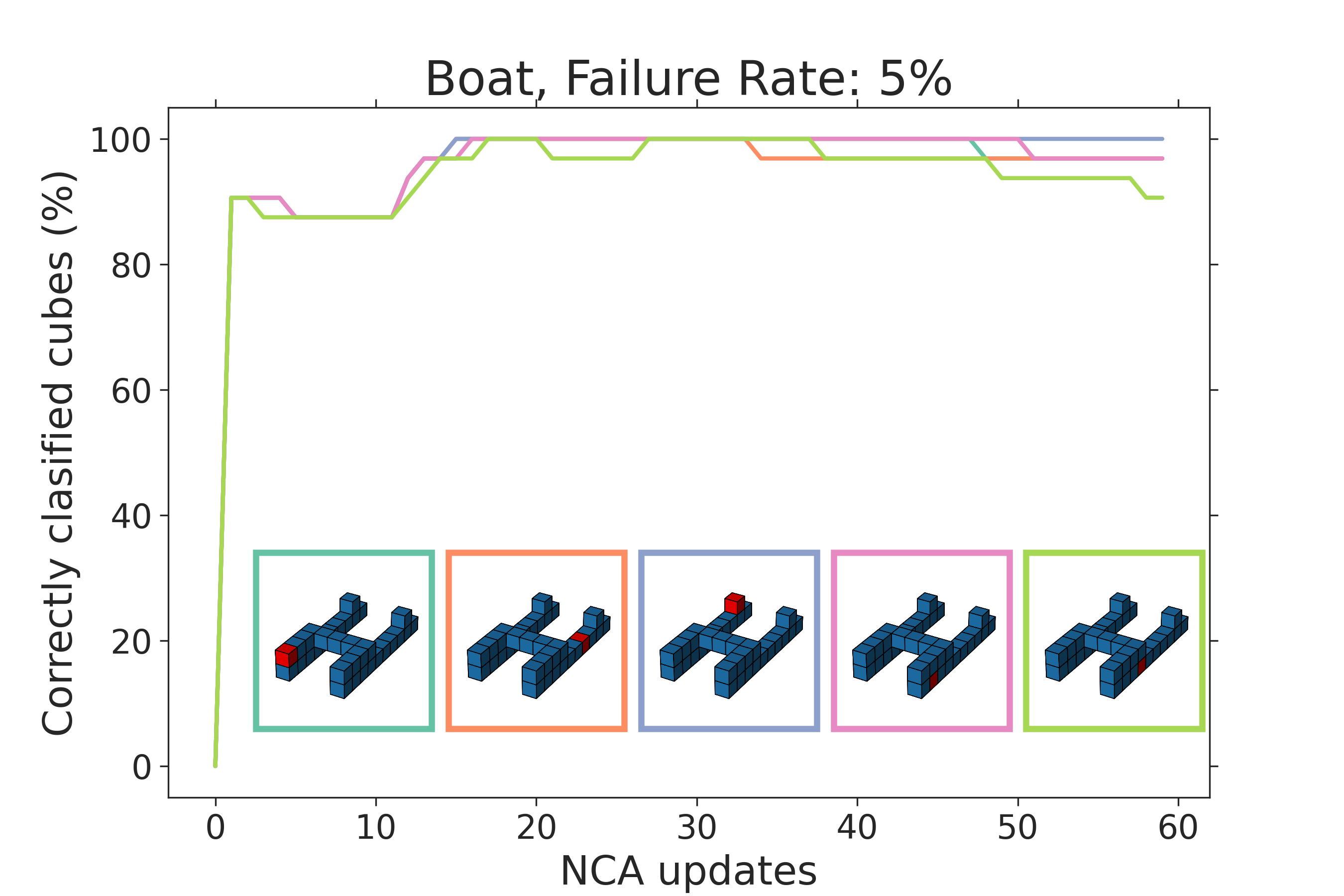}
  \includegraphics[width=0.3\textwidth]{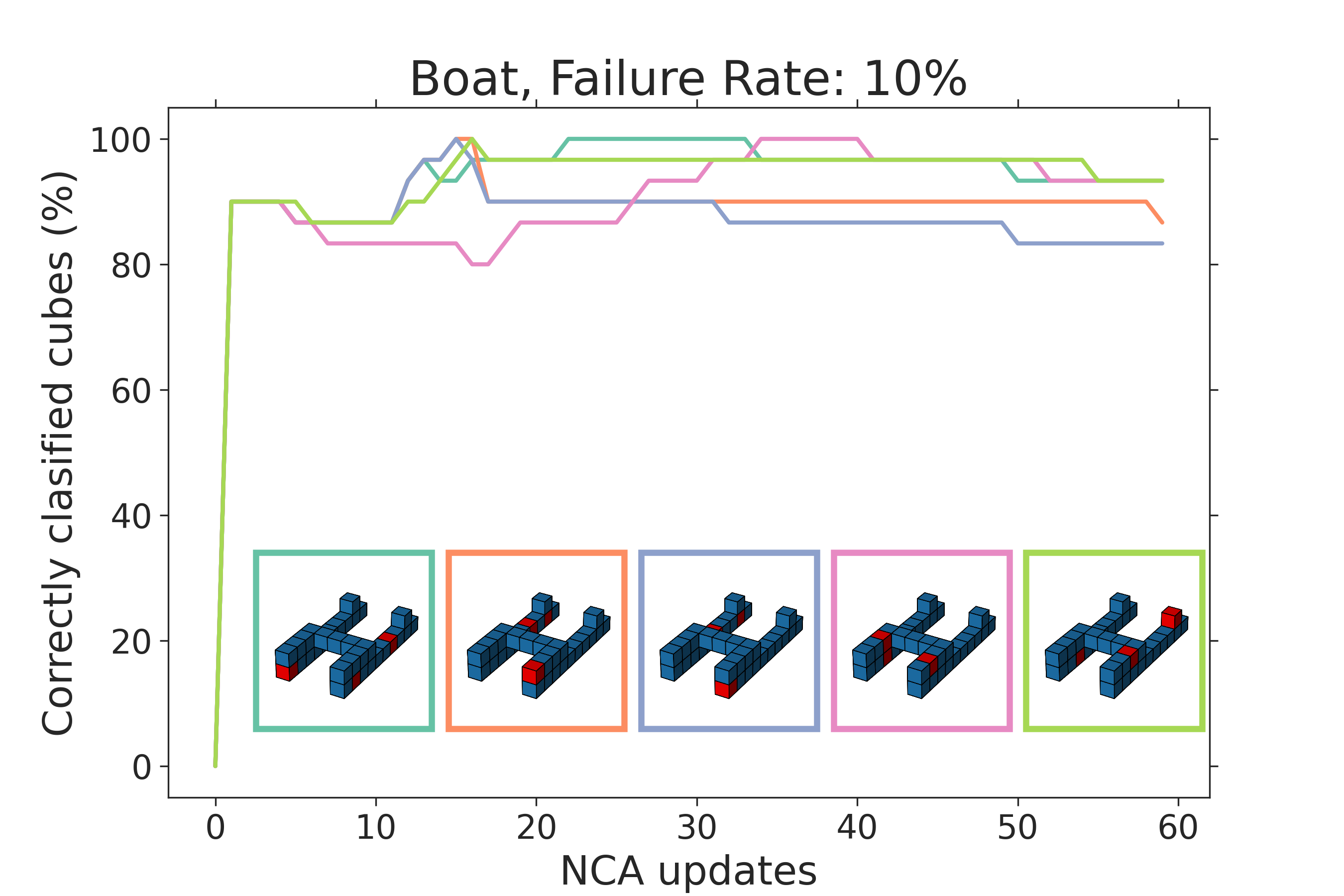}
  \includegraphics[width=0.3\textwidth]{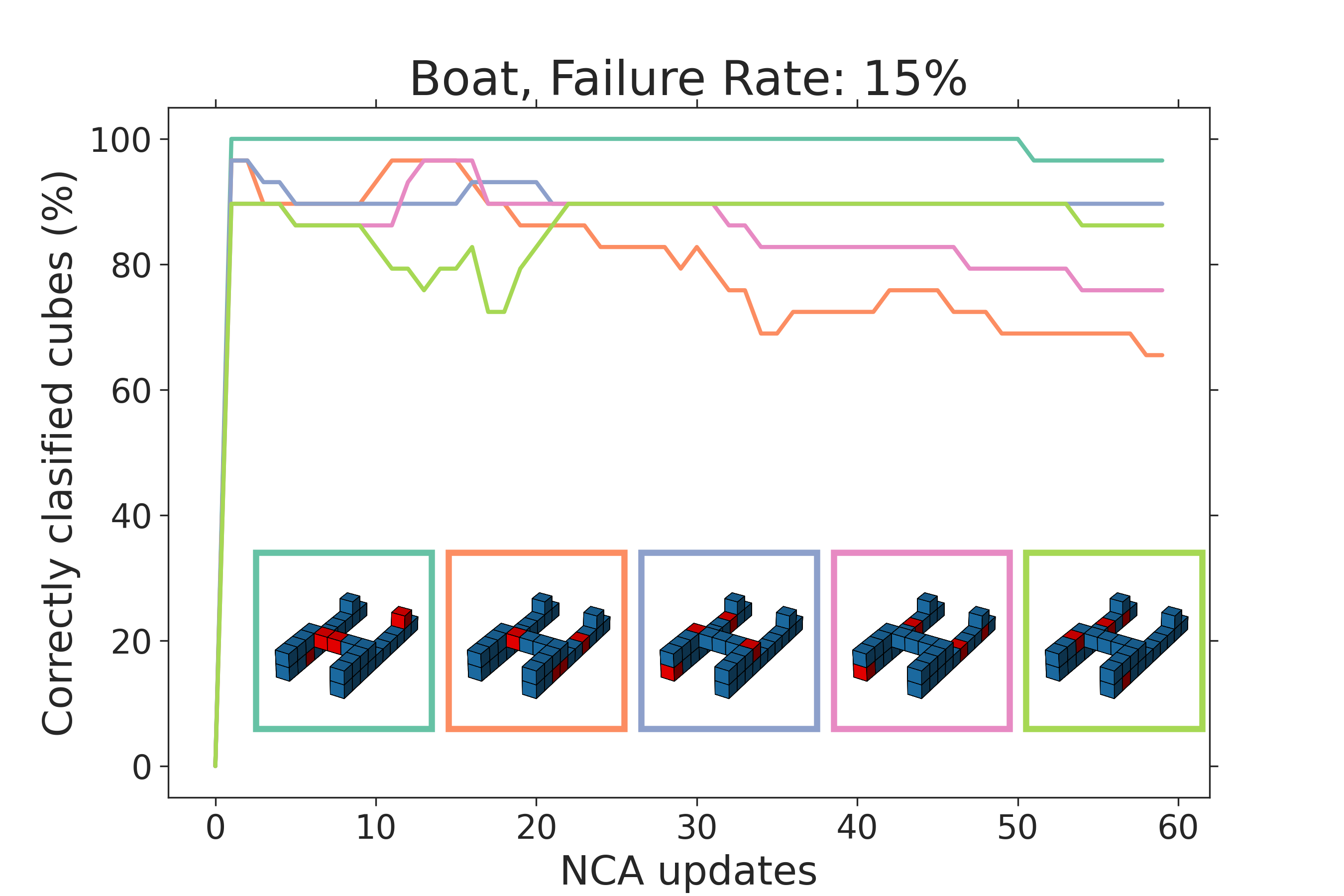}
  \caption{\textbf{Classification experiments for the boat in hardware}. \normalfont The four graphs show the percentage of correctly classified cubes for failure rates of 0, 5, 10 and 15\%. The damaged modules for each experiment are indicated with small icons inside the graphs.} 
    \label{fig:table_failure_rate_boat}
\end{figure}

\begin{figure}[H]
  \centering
  \includegraphics[width=0.3\textwidth]{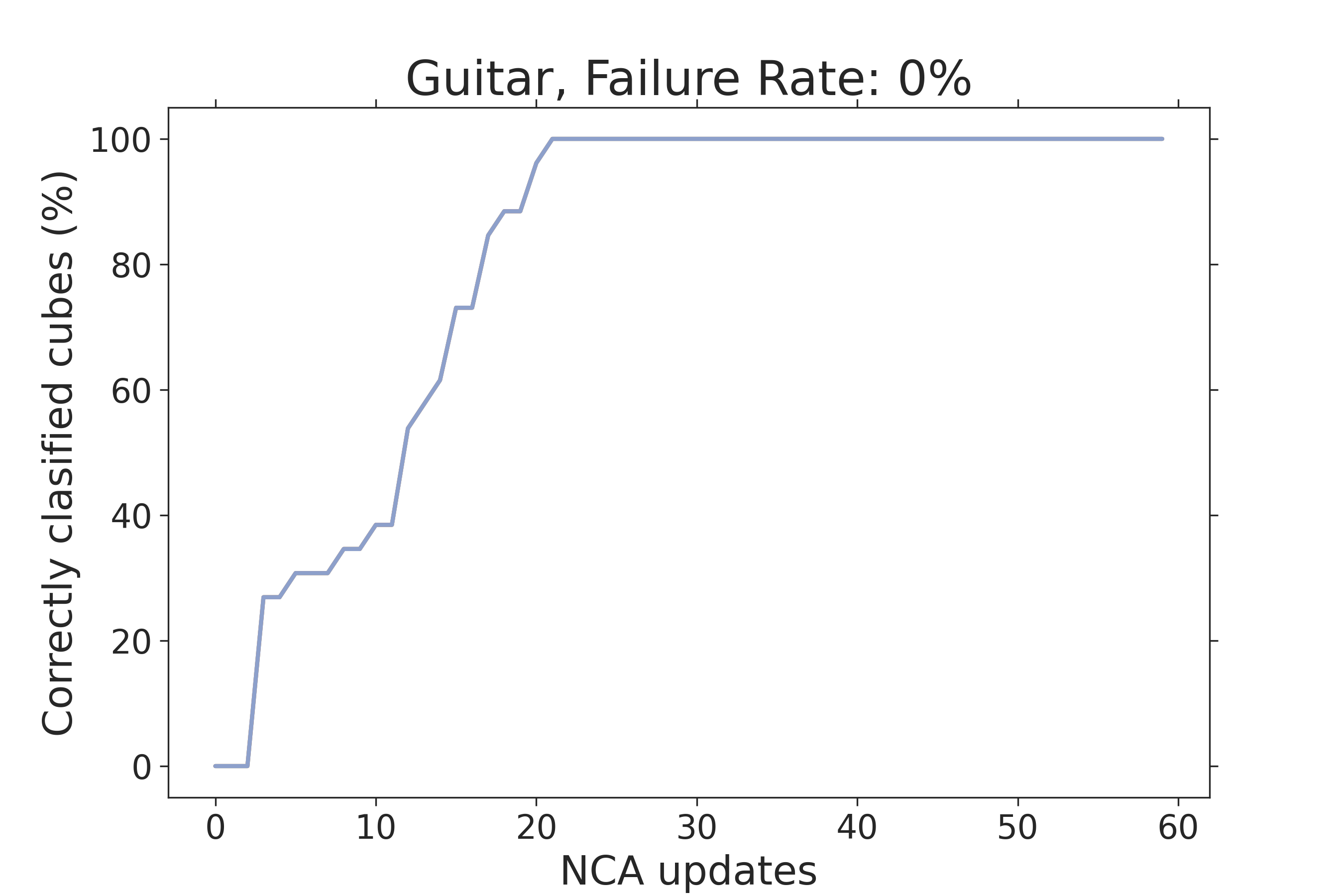}
  \includegraphics[width=0.3\textwidth]{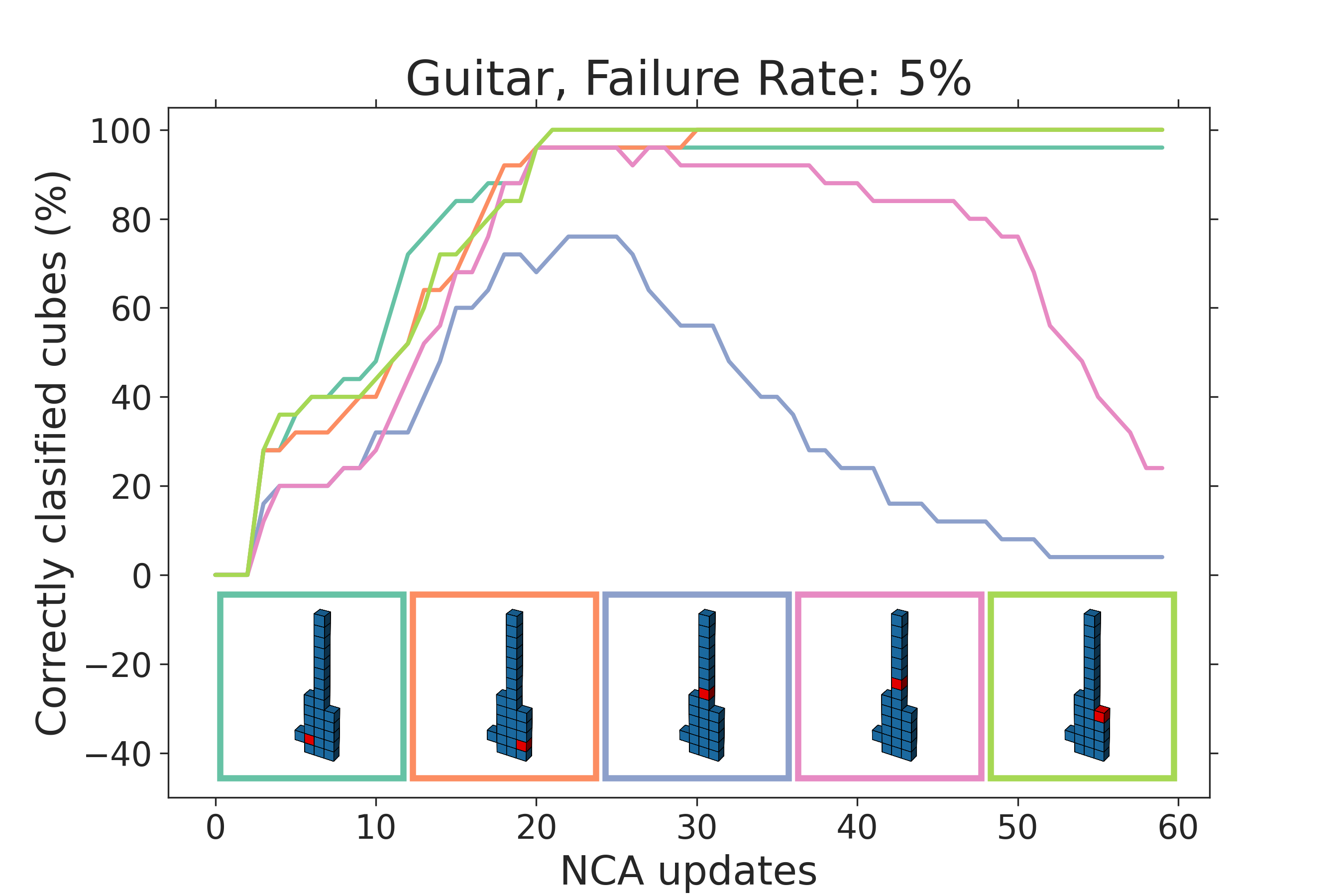}
  \includegraphics[width=0.3\textwidth]{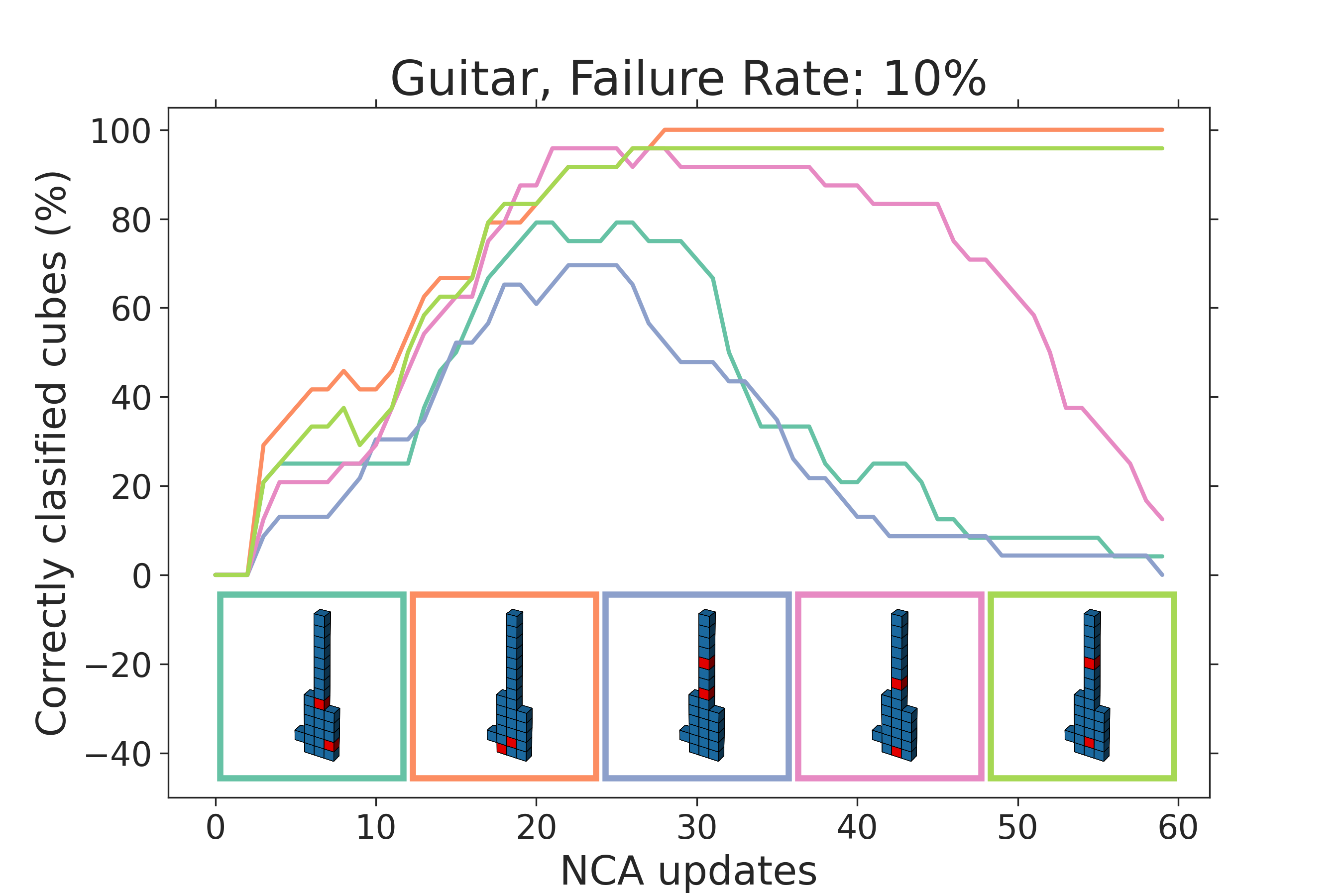}
  \includegraphics[width=0.3\textwidth]{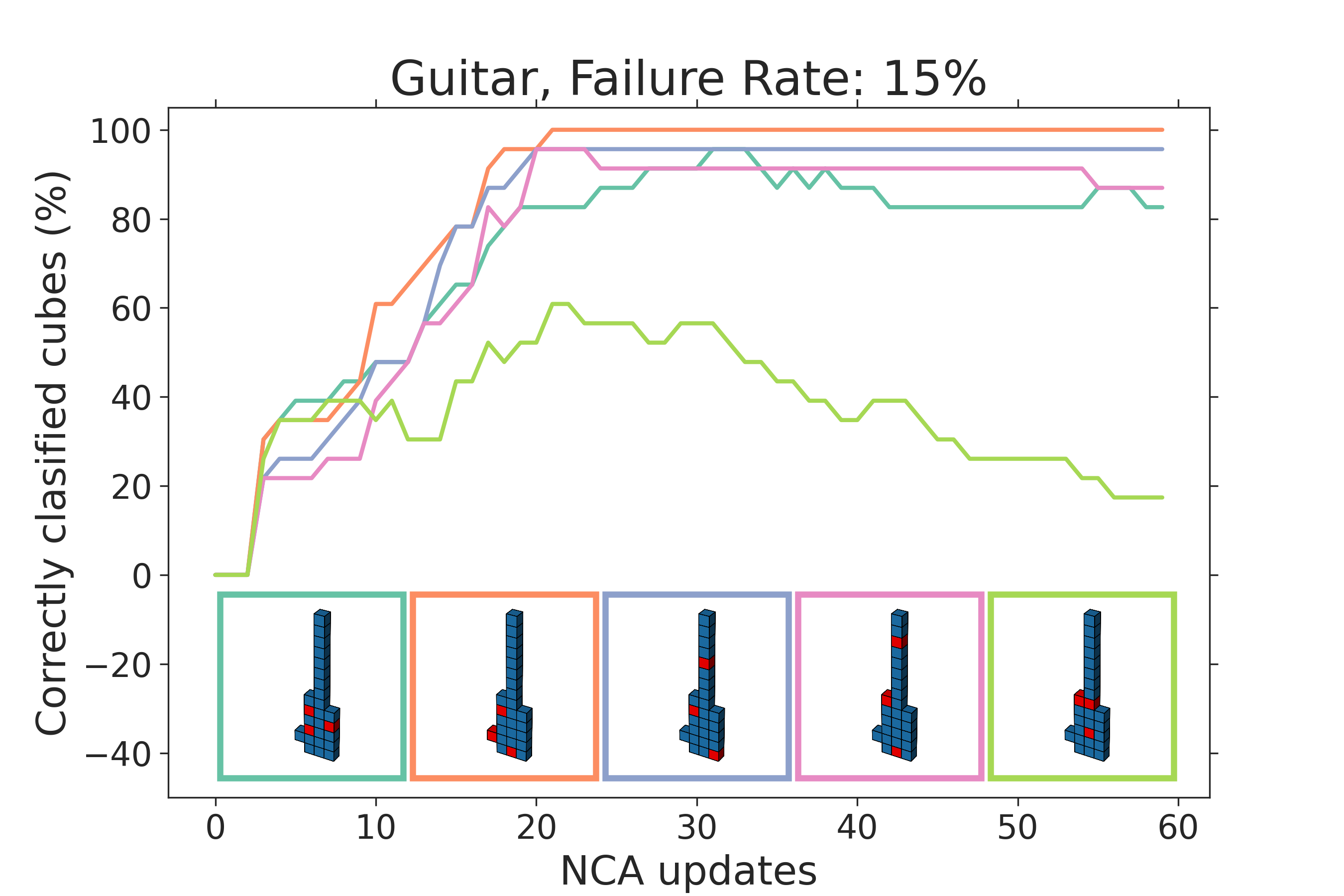}
  \caption{\textbf{Classification experiments for the guitar in hardware}. \normalfont The four graphs show the percentage of correctly classified cubes for failure rates of 0, 5, 10 and 15\%. The damaged modules for each experiment are indicated with small icons inside the graphs.} 
    \label{fig:table_failure_rate_guitar}
\end{figure}

Figs.~\ref{fig:table_failure_rate_table}, \ref{fig:table_failure_rate_plane}, \ref{fig:table_failure_rate_boat}, and \ref{fig:table_failure_rate_guitar}   show more detailed hardware results for a variety of shapes under varying levels of faulty cells. 

\section*{Electronic Supplementary Information}
\begin{itemize}
    \item \textbf{Movie S1.}  Video of four different objects (table, boat, plane, guitar) during self-classification. The process takes around three minutes in real-time, with all the cubes reaching the correct classification.  
    \item \textbf{Movie S2.}  Video shows robustness to different levels of faulty cell (ranging from 5\% -- 15\%) for different objects. 
    \item \textbf{Movie S3.} Video shows out-of-distribution generalization to shapes not seen during training, such as modified table, and scaled-down versions of a plane, a guitar, a round table. 
    \item \textbf{Movie S4.} Video of the different hidden channels during self-classification of different shapes. 
\end{itemize}

\section*{Acknowledgments}
 The authors thank Scarlett Fantasia Avalon Petersen, Thor O'Hagan Petersen, Andreas Dehn, and Theodor Christian Kier for helping with the assembly of the bricks.
 
\textbf{Funding:} This work was supported by the European Union (ERC, GROW-AI, 101045094) and Novo Nordisk Foundation Synergy Grant (``REPROGRAM'').

\textbf{Competing interests:} The authors declare that they have no competing interests.

\newpage
\bibliography{sn-bibliography}

\end{document}